\documentclass[twocolumn]{autart}    
\usepackage{amsmath}
\usepackage{amssymb}
\usepackage{nccmath}
\usepackage{graphicx}          
\usepackage{outlines}
\usepackage{multirow}
\usepackage{xcolor}
\usepackage{subcaption}
\usepackage{array}
\usepackage[normalem]{ulem}

\begin{document}
	\setlength{\abovedisplayskip}{3pt}
	\setlength{\belowdisplayskip}{3pt}

\begin{frontmatter}

\title{An Analytic Layer-wise Deep Learning Framework with Applications to Robotics} 

\thanks[footnoteinfo]{{This work was supported by the Agency For Science, Technology and Research of Singapore (A*STAR), under the AME Individual Research Grants 2017 (Ref. A1883c0008). The material in this paper was not presented at any IFAC meeting.} \\
Corresponding author is C.C. Cheah.}

\author[NTU]{Huu-Thiet Nguyen},    
\author[NTU]{Chien Chern Cheah}\ead{ECCCheah@ntu.edu.sg},
\author[YU]{Kar-Ann Toh}

\address[NTU]{School of Electrical and Electronic Engineering, Nanyang Technological University, 50 Nanyang Avenue, Singapore 639798}  
\address[YU]{School of Electrical and Electronic Engineering, Yonsei University, Seoul, Korea 03722}             


\begin{abstract}                          
Deep learning (DL) has achieved great success in many applications, but it has been less well analyzed from the theoretical perspective. {The unexplainable success of black-box DL models has raised questions among scientists and promoted the emergence of the field of explainable artificial intelligence (XAI). In robotics, it is particularly important to deploy DL algorithms in a predictable and stable manner} as robots are active agents that need to interact safely with the physical world. This paper presents an analytic deep learning framework for fully connected neural networks, which can be applied for both regression problems and classification problems. Examples for regression and classification problems include online robot control and robot vision. We present two layer-wise learning algorithms such that the convergence of the learning systems can be analyzed. Firstly, an inverse layer-wise learning algorithm for multilayer networks with convergence analysis for each layer is presented to understand the problems of layer-wise deep learning. Secondly, a forward progressive learning algorithm where the deep networks are built progressively by using single hidden layer networks is developed to achieve better accuracy. It is shown that the progressive learning method can be used for fine-tuning of weights from convergence point of view. The effectiveness of the proposed framework is illustrated based on classical benchmark recognition tasks using the MNIST and CIFAR-10 datasets and the results show a good balance between performance and explainability.  The proposed method is subsequently applied for online learning of robot kinematics and experimental results on kinematic control of UR5e robot with unknown model are presented. 
\end{abstract}

\end{frontmatter}

\section{Introduction}
{Artificial} neural networks (ANNs), or simply neural networks (NNs), have been widely deployed in data related problems, such as regression analysis, data processing, classification, control and robotics. In robotic applications, the use of ANNs can be traced back to late 1980s \cite{miller1989real}. The ANNs have been proved to be universal approximators \cite{hornik1989multilayer,Poggio201907369} where their great potential in identification and control of dynamic systems was discussed in \cite{narendra1990identification}. Hence, they have been regarded as a potential approach to deal with nonlinearities, modeling uncertainties and disturbances in robot control systems. \vspace{-2mm} 

For years, many results in feedback control of robots have been obtained by focusing on regression problems based on \textit{simple shallow networks}. Most studies are based on one hidden layer networks such as the single-hidden layer feedforward networks (SLFNs) and radial basis function networks (RBFNs). There are two notable approaches in learning of these networks in robotics: the first approach is training only the output weights, which is still popular until recently \cite{sanner1992gaussian,yang2014neural,li2014adaptive,he2016adaptive,yang2016neural,licitra2019single,qiu2019command,mu2019learning,yang2019force,lyu2020data}; the second approach focuses on training both input and output weights of the networks \cite{chen1995adaptive,lewis1996multilayer,fierro1998control,cheng2009adaptive,chu2016observer}. \vspace{-2mm}

\textit{In the first approach} of training only the output weights, by using linear output activation functions, the algorithms for updating the last layer of weights of these networks resemble those adaptive control techniques where the {model} is linear with tunable parameters. Similar to adaptive control, the convergence and stability of these algorithms can be ensured by using the Lyapunov method. Among those early studies in NN based control, Sanner and Slotine \cite{sanner1992gaussian} analyzed the approximation capability of Gaussian networks and employed them in control of systems with dynamic uncertainties. The RBFNs were employed in an indirect controller of a subsystem and in an adaptive NN model reference controller of another subsystem in underactuated wheeled inverted pendulums \cite{yang2014neural}. While robot control problems are usually formulated for trajectory tracking task, a region adaptive NN controller with a unified objective bound was synthesized for robot control in task space \cite{li2014adaptive}. In \cite{he2016adaptive}, He et al. proposed an adaptive NN control technique for robots with full-state constraints which guaranteed the asymptotic tracking. Global stability was ensured by using a NN-based controller for dual-arm robots with dynamic uncertainties \cite{yang2016neural}. {Recently, the approach {of training only the output weights} has been adopted for indirect herding \cite{licitra2019single}, MIMO nonlinear systems with full-state constraints \cite{qiu2019command}, quadrotors with time-varying and coupling uncertainties \cite{mu2019learning}, and teleoperation control systems \cite{yang2019force}}. For robots with unknown Jacobian matrix, an SLFN-based controller that guaranteed the stability of the system were proposed in \cite{lyu2020data}. \vspace{-2mm}

\textit{In the second approach} of training weights in both layers, besides the output weights, the input weights of the network are also adjusted. In the 1990s, Chen and Khalil \cite{chen1995adaptive} provided convergence analysis of a learning algorithm that was based on the backpropagation (BP) and gradient descent (GD) in multilayer NN control of nonlinear systems. In 1996, Lewis et al. \cite{lewis1996multilayer} proposed a learning algorithm for updating 2 layers of weights (input weights and output weights) in a SLFN with Lyapunov-based convergence analysis. The study paved the way for more research works in applying SLFNs in dynamic systems like robotics. In \cite{fierro1998control}, control of nonholonomic mobile robots was studied and a NN controller was introduced to deal with disturbances and unmodeled dynamics. In \cite{cheng2009adaptive}, an adaptive NN-based controller was  proposed for manipulators with kinematic and dynamic uncertainties. {A recurrent NN based control was also developed for remotely operated vehicles in \cite{chu2016observer}.} {In general, these works have focused on dynamic systems (of the form \(\dot{\boldsymbol{z}}=\boldsymbol{f}(\boldsymbol{z})\) where \(\boldsymbol{z}\) is the state variable) with the use of shallow networks with linear output activation functions. For general input-output mappings of the form \(\boldsymbol{y} = \boldsymbol{f}(\boldsymbol{z})\) which include both regression and classification problems, deep networks have been shown to demonstrate better properties as compared to the shallow counterparts \cite{bengio2007scaling,bengio2009learning,srivastava2015training,mhaskar2017and}. \vspace{-2mm}

Deep networks \cite{bishop1995neural,glorot2011deep} have become dominant in machine learning (ML) applications in the last decade. {Comparing with} shallow networks, deep architectures have been shown to be more efficient in terms of number of tunable parameters \cite{bengio2007scaling,larochelle2007empirical}. They are also believed to be better {for learning} generalization \cite{srivastava2015training,poggio2017and}.} The deep NNs have also been shown to be more powerful in function approximation and classification than the single hidden layer networks \cite{mhaskar2016deep}. The current boom of ML applications in many aspects of our life is greatly attributed to deep learning (DL) algorithms \cite{goodfellow2016deep,schmidhuber2015deep} in which backpropagation (BP) \cite{rumelhart1986learning} plays a major role. DL has replaced many conventional learning algorithms which saw disadvantages in processing raw data \cite{lecun2015deep}. Many unprecedented successes in image recognition have been achieved by the convolutional neural networks (CNNs) \cite{lecun2015deep,krizhevsky2012imagenet}. 
However, DL has been less well analyzed from theoretical perspective and DL models remain difficult to understand despite the tremendous successes \cite{sejnowski2020unreasonable}. \vspace{-2mm}

Various attempts have been made to understand the properties of deep networks. Layer-wise learning {\cite{mostafa2018deep,marquez2018deep,nokland2019training}} is one of the methods to dissect a large network into smaller pieces. One method of training network layer-by-layer is using matrix pseudoinverse as in \cite{guo2004pseudoinverse} and together with functional inverse, as developed in \cite{toh2018analytic}. The method \cite{toh2018analytic} does not require any computation of the gradient information and can be applied to both regression and classification problems. However, its performance still lags behind the state-of-art gradient descent DL algorithms in many applications. Employing the layer-wise method in \cite{toh2018analytic}, an iterative learning algorithm for offline regression problem of robot kinematics was developed \cite{nguyen2019data} but it was again limited to shallow networks with one or two hidden layers. The algorithm \cite{nguyen2019data} was built and analyzed in continuous time and hence could not be generalized for classification problems. In addition, the learning of the input layer was ignored and the weights obtained were time-varying which thus required averaging of the weights. Another approach for training deep networks layer-wisely is greedy layer-wise learning which can be found in network pre-training \cite{bengio2007greedy,erhan2010does} and forward thinking algorithm \cite{hettinger2017forward}. In this methodology \cite{bengio2007greedy,erhan2010does,hettinger2017forward}, training of the multilayer NN is performed by adding the layers in a forward manner, starting from the first layer based on the training of shallow networks. One hidden layer is added at a time, and each training step involves a single hidden layer feedforward network. After training each SLFN, the input weights are kept while the output weights are discarded. A new hidden layer is then added on top of the hidden layer of the last SLFN to create a new SLFN. However, despite the good performance especially in classification problems, there is no convergence analysis for these algorithms \cite{bengio2007greedy,erhan2010does,hettinger2017forward}, which is a common problem in the ML literature. \vspace{-2mm}

Robots are active agents which interact physically with the real world, and applying DL tools in robot control {requires} careful consideration \cite{sunderhauf2018limits}. When employing a deep network in control of robotic systems, one should guarantee the stability, convergence and robustness because robots need to be operated in a safe and predictable manner. Most DL algorithms used in ML community lack theoretical supports for convergence analysis. Therefore, in spite of many elegant results from ML research, very few can be directly used in the area of robot control for the reason of safety. Recently, there has been an increasing need of building interpretable and understandable deep neural networks \cite{montavon2018methods}. As a result, the field of explainable artificial intelligence (XAI) has begun to attract more attention from academics \cite{gunning2019xai}. An XAI project launched by DARPA has aimed for developing accountable AI with a compromise between performance and explainability \cite{gunning2017explainable}. Therefore, establishing a reliable theoretical framework for constructing and training deep networks, which ensures the convergence, could open up many XAI applications to robotics.  \vspace{-2mm}

This paper aims to develop a theoretical framework for multilayer NNs which can be efficiently applied in operations of robotic systems. Our main focus is on the study of learning algorithms and training methodologies for multilayer NNs to {enable them for deployment} in a reliable and explainable manner, which is desirable for control of active agents such as robots. {Unlike most literature on machine learning where the algorithms are considered from optimization perspective, our methodology is formulated based on Lyapunov-like analysis which aims to bridge the gaps between robotics, control and machine learning. In this paper}, an inverse learning algorithm is first formulated to understand the issues and difficulties of establishing analytic layer-wise deep learning and based on this study, an analytic forward progressive algorithm is then proposed to overcome the problems. The main contributions of this paper are: \vspace{-3mm}
\begin{itemize}
	\item[i,] Development of a theoretical framework to ensure the convergence of the layer-wise deep learning algorithms. To the best of our knowledge, there is currently no theoretical result for analysis of deep learning of fully connected neural networks to ensure convergence for safe operation of robotic systems.
	\item[ii,] Development of a forward progressive layer-wise learning algorithm for deep networks in which the general input-output function \(\boldsymbol{y} = \boldsymbol{f}(\boldsymbol{z}) \) is also considered. Based on the convergence analysis, it is shown that the proposed algorithm can be used for fine-tuning of weights.
	\item[iii,] Development of a systematic learning or training methodology in which deep networks can be built gradually for reliable operations in both online and offline robotic applications. 
	\vspace{-2mm}
\end{itemize}
The proposed framework is applied to two recognition tasks using the MNIST {\cite{lecun1998gradient,lecun1998mnist}} and CIFAR-10 {\cite{krizhevsky2009learning,krizhevsky2014cifar}} databases and an online kinematic robot control task using a UR5e manipulator. Experimental results are presented to illustrate the performance of the proposed algorithm. It is shown that forward progressive learning can achieve similar accuracy as compared to gradient descent method but the main advantage is that the convergence of proposed algorithm can be established in a systematic way. \vspace{-2mm}

\section{Problem Statement} \label{sec: problm_state} \vspace{-2mm}

Consider a mapping between the input variable \(\boldsymbol{z} \in \mathbb{R}^m\) and the output variable \(\boldsymbol{y} \in \mathbb{R}^p\)
\begin{flalign}
\boldsymbol{y} = \boldsymbol{f}(\boldsymbol{z}) && \label{eq:1}
\vspace{-2mm}
\end{flalign}
The function \(\boldsymbol{f}: \mathbb{R}^m \to \mathbb{R}^p\) is assumed to be unknown, but can be approximated by available input and output (target) data which are referred to as training data. 
Our objective is to develop a theoretical framework to achieve an approximation (model) of the function \(\boldsymbol{f}\) based on the training data, so that this model can predict well on unknown new data. \vspace{-2.5mm}

Based on the output variable \(\boldsymbol{y}\), the problem can be divided into two main types: \vspace{-3mm}
\begin{itemize}
	\item When \(\boldsymbol{y}\) is a continuous variable, the problem is known as a regression problem.
	\item When \(\boldsymbol{y}\) is a categorical variable, the problem is known as a classification problem.  \vspace{-3mm}
\end{itemize}
In the area of robotics, both types of problems can be found. For instance, when a robot needs to identify (and label) the objects within its work space, a classification or a recognition task should be done; but how the robot makes movement by rotating its joints to reach the position of the object would be a regression problem. \vspace{-2.5mm}

In order to approximate \(\boldsymbol{f}\), a multilayer feedforward neural network (MLFN) is used. In this paper, we present two techniques for training the MLFNs. The first one is called \textit{inverse layer-wise learning} which is presented in section \ref{sec: main_1}. The second one is called \textit{forward progressive learning} which is presented in section \ref{sec: main_2}. \vspace{-2.5mm}

\begin{figure}[ht]
	\centering
	\includegraphics[width=3.1in]{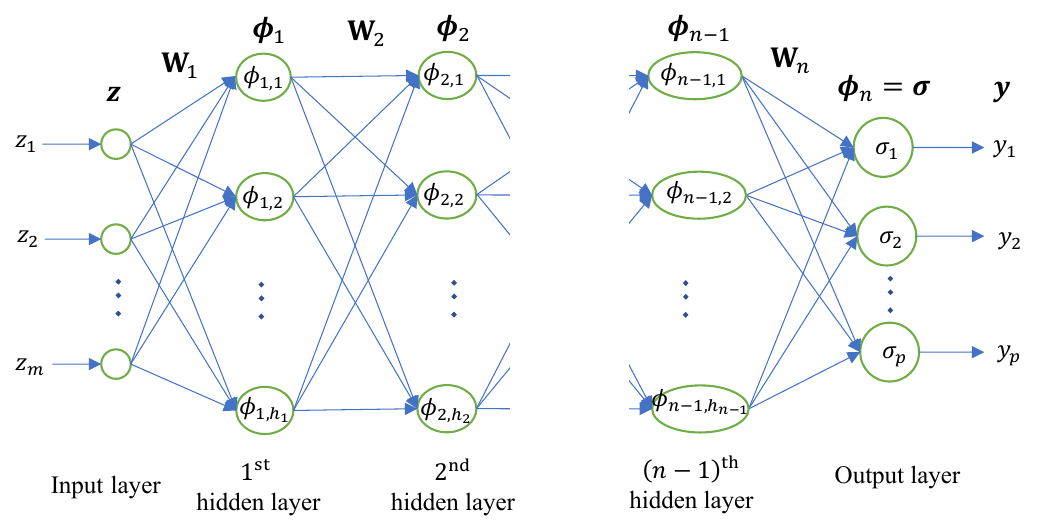}
	\caption{Structure of an \(n\)-layer feedforward neural network with input $\boldsymbol{z}$ and target output $\boldsymbol{y}$.}
	\label{fig:mlfn}
\end{figure}

An illustration of an \(n\)-layer MLFN (\(n-1\) hidden layers) is shown in Fig. \ref{fig:mlfn}. In this MLFN, the number of hidden units for the \(j^{\text{th}}\) hidden layer \((1\leq j \leq n-1)\) is denoted as \(h_j\); the activation functions for the \(j^{\text{th}}\) hidden layer are denoted as \(\boldsymbol{\phi}_j\): \(\boldsymbol{\phi}_j=[\phi_{j,1},\phi_{j,2},\ldots,\phi_{j,h_j}]^T\); the activation functions for output layer are denoted as \(\boldsymbol{\phi}_{n}\): \(\boldsymbol{\phi}_{n}=\boldsymbol{\sigma}=[\sigma_{i,1},\sigma_{i,2},\ldots,\sigma_{i,p}]^T\) (and hence \(h_{n} \triangleq p\)); The \(n\) weight matrices are denoted (from input layer to output layer) as \(\mathbf{W}_1, \mathbf{W}_2, \dots, \mathbf{W}_{n}\) where \(\mathbf{W}_1\) is the matrix of input weights and \(\mathbf{W}_{n}\) is the matrix of output weights. 
The output of the MLFN as shown in Fig. \ref{fig:mlfn} is given as follows
\begin{eqnarray}
\boldsymbol{y}_{\mathcal{NN}} &=& \boldsymbol{\phi}_{n}(\mathbf{W}_n \boldsymbol{\phi}_{n-1}(\mathbf{W}_{n-1} \boldsymbol{\phi}_{n-2}( \dots\nonumber\\
&&{\;\;\;\;\;\;\;\;\;} {\dots} \mathbf{W}_3 \boldsymbol{\phi}_2(\mathbf{W}_2 \boldsymbol{\phi}_1(\mathbf{W}_1 \boldsymbol{z}))\dots))) \label{eq:mlfn_output}
\end{eqnarray}
\noindent Since the weights are the tunable parameters, the output of the MLFN in \eqref{eq:mlfn_output} can be written as
\begin{flalign}
\boldsymbol{y}_{\mathcal{NN}} = \boldsymbol{y}_{\mathcal{NN}}(\mathbf{W}_j|_{j=1}^n,\boldsymbol{z}) && \label{eq:mlfn_output_short}
\end{flalign}
The denotation shows the dependence of the network output on the weights \(\mathbf{W}_j\) \((j = 1, \ldots, n)\) and the input variable \(\boldsymbol{z}\).
The output after the \({(j-1)}^\text{th}\) hidden layer (activation values after \(\boldsymbol{\phi}_{j-1}\)) is given as
\begin{flalign}
\boldsymbol{y}_{\mathcal{NN}{(j-1)}}(\mathbf{W}_l&|_{l=1}^{j-1},\boldsymbol{z})=  \boldsymbol{\phi}_{j-1}(\mathbf{W}_{j-1} \boldsymbol{\phi}_{j-2}(\dots& \nonumber \\
& \dots \mathbf{W}_3 \boldsymbol{\phi}_2(\mathbf{W}_2 \boldsymbol{\phi}_1(\mathbf{W}_1 \boldsymbol{z}))\dots))& \label{eq:mlfn_output_j-1} 
\end{flalign}
\noindent{\textbf{Remark 1:} Proving the convergence of deep networks is a difficult {but important task} since deep networks are well-known to be high-dimensional, and the optimization problem is highly non-convex. Majority of ML scientists are applying DL in attempts to push empirical results to {higher peaks} for {various} ML tasks despite the fact that the {underlying} reason behind the success of DL remains {largely} unclear \cite{sejnowski2020unreasonable,gunning2019xai}. In this paper, the convergence problem of fully connected deep NNs is formulated and solved by developing a layer-wise learning framework {where the deep NNs are layer-wisely built and trained through learning of the basic} two-layer {structures}. {Such development would} widen the potential applications of deep learning {particularly when} a safe and predictable {outcome is desired}.}\vspace{-2.5mm}
\section{Inverse Layer-wise Learning of Multilayer Feedforward Neural Networks} \label{sec: main_1} \vspace{-2.5mm}

In this approach, the MLFN is trained layer-by-layer to ensure the convergence, which means one layer of weights is learned at a time. 
In \cite{toh2018analytic}, by using functional inverse and matrix pseudoinverse, equation \eqref{eq:mlfn_output} has been treated as follows with \(\boldsymbol{y}_{\mathcal{NN}} = \boldsymbol{y}\)
\begin{flalign}
&{} \boldsymbol{y} = \boldsymbol{\phi}_{n}(\mathbf{W}_n \boldsymbol{\phi}_{n-1}( \mathbf{W}_{n-1} \boldsymbol{\phi}_{n-2}( &\nonumber\\
&{\;\;\;\;\;\;\;\;\;\;\;\;\;\;\;\;\;\;\;\;\;\;\;\;\;\;}  \dots \mathbf{W}_3 \boldsymbol{\phi}_2(\mathbf{W}_2 \boldsymbol{\phi}_1(\mathbf{W}_1 \boldsymbol{z}))\dots)))& \label{eq:learn_wn}\\
&\to \mathbf{W}_n^{\dagger}\boldsymbol{\phi}_{n}^{-1}(\boldsymbol{y})=\boldsymbol{\phi}_{n-1}(\mathbf{W}_{n-1} \boldsymbol{\phi}_{n-2}(&\nonumber\\
&{\;\;\;\;\;\;\;\;\;\;\;\;\;\;\;\;\;\;\;\;\;\;\;\;\;\;}\dots \mathbf{W}_3 \boldsymbol{\phi}_2(\mathbf{W}_2 \boldsymbol{\phi}_1(\mathbf{W}_1 \boldsymbol{z}))\dots))&\label{eq:learn_wn-1}\\ 
&\to{\;}\cdots& \nonumber \\
&\to \mathbf{W}_2^{\dagger}\boldsymbol{\phi}_2^{-1}( \cdots  \mathbf{W}_{n-1}^{\dagger}\boldsymbol{\phi}_{n-1}^{-1}(\mathbf{W}_n^{\dagger}\boldsymbol{\phi}_{n}^{-1}(\boldsymbol{y}))\cdots) &\nonumber\\
&{\;\;\;\;\;}= \boldsymbol{\phi}_1(\mathbf{W}_1 \boldsymbol{z}) &\label{eq:learn_w0}
\vspace{-3mm}
\end{flalign}
where \(\mathbf{W}_n^{\dagger}, \mathbf{W}_{n-1}^{\dagger}, \ldots, \mathbf{W}_2^{\dagger}\) are the Moore-Penrose inverses (or pseudoinverses) of matrices  \(\mathbf{W}_n, \mathbf{W}_{n-1}, \ldots\), \(\mathbf{W}_2\) respectively; \(\boldsymbol{\phi}_{n}^{-1}, \boldsymbol{\phi}_{n-1}^{-1}, \ldots, \boldsymbol{\phi}_2^{-1}\) are vectors of inverse functions of respective \(\boldsymbol{\phi}_{n}, \boldsymbol{\phi}_{n-1}, \ldots, \boldsymbol{\phi}_2\). \vspace{-3mm}

The inverse layer-wise learning is conducted through two stages in sequence: backward and forward. The learning process starts with the backward stage (subsection \ref{sec:backward}), where the MLFN is trained layer-wisely from the output layer to the input layer. After that, the forward stage takes place (subsection \ref{sec:forward}), where the network is trained layer-wisely in the forward direction, from the input layer to the output layer. 
{Unlike in \cite{toh2018analytic} where the kernel and range space to solve linear equations {were used}, we propose nonlinear update laws to find the weight matrices incrementally so that the convergence is ensured.} \vspace{-2.5mm}

\subsection{Backward Stage of Inverse Layer-wise Learning} \label{sec:backward}
\vspace{-2.5mm}
In the first stage of inverse layer-wise learning, the network is trained layer-by-layer from \(\mathbf{W}_n\) to \(\mathbf{W}_1\). That is, \(\mathbf{W}_n\) is trained first. Then comes \(\mathbf{W}_{n-1}, \mathbf{W}_{n-2}, \ldots\). The backward stage ends with the learning of the input weights \(\mathbf{W}_1\). Let us now look into the details of how the  weights in each layer are trained, starting from \(\mathbf{W}_n\). Prior to that, all the weight matrices \(\mathbf{W}_1, \mathbf{W}_2, \dots, \mathbf{W}_{n-1}, \mathbf{W}_{n}\) are first randomly initialized. Let \(\bar{\mathbf{W}}_1^{\star}, \bar{\mathbf{W}}_2^{\star}, \dots, \bar{\mathbf{W}}_{n-1}^{\star}, \bar{\mathbf{W}}_{n}^{\star}\) be the respective values of these matrices after initialization. \vspace{-2.5mm}

\subsubsection{Learning of the output weights \(\mathbf{W}_n\)} \vspace{-2.5mm}
During learning of \(\mathbf{W}_n\) (or the \(n^\text{th}\) layer), the other weight matrices are frozen at their initialized values \(\bar{\mathbf{W}}_1^{\star}, \bar{\mathbf{W}}_2^{\star}, \dots, \bar{\mathbf{W}}_{n-1}^{\star}\). Using these values, we can compute the input of the \(n^\text{th}\) layer, which is also the output after the \({(n-1)}^\text{th}\) hidden layer, by using \eqref{eq:mlfn_output_j-1}
\begin{flalign}
&\bar{\boldsymbol{\phi}}_{n-1}^{\star} = \boldsymbol{\phi}_{n-1}(\bar{\mathbf{W}}_{n-1}^{\star}\boldsymbol{\phi}_{n-2}(\dots  \boldsymbol{\phi}_2(\bar{\mathbf{W}}_2^{\star} \boldsymbol{\phi}_1(\bar{\mathbf{W}}_1^{\star} \boldsymbol{z}))\dots))& \label{eq:input_wn}  \vspace*{-3mm} 
\end{flalign}
Hence, the output of this \(n^\text{th}\) layer (and also of the whole MLFN), denoted as \(\boldsymbol{y}_{\mathcal{NN}n}\), is given as 
\begin{flalign}
&\boldsymbol{y}_{\mathcal{NN}n}(\mathbf{W}_n,\bar{\boldsymbol{\phi}}_{n-1}^\star)
= \boldsymbol{\phi}_{n}({\mathbf{W}}_{n} \bar{\boldsymbol{\phi}}_{n-1}^\star)&\label{eq:mlfn_wn_out} \vspace*{-3mm} 
\end{flalign} 
This is actually equal to the right-hand side of \eqref{eq:learn_wn} when setting \(\mathbf{W}_1, \ldots, \mathbf{W}_{n-1}\) as \(\bar{\mathbf{W}}_1^{\star}, \ldots, \bar{\mathbf{W}}_{n-1}^{\star}\), respectively. Hence, the target for learning of \(\mathbf{W}_{n}\), denoted as \(\boldsymbol{y}_{n}\), is the direct target \(\boldsymbol{y}\) of the whole MLFN as seen on the left-hand side of \eqref{eq:learn_wn}. That is
\begin{flalign}
&\boldsymbol{y}_n = \boldsymbol{y} &\label{eq:target_wn}
\vspace{-2mm}
\end{flalign}
Given \(\bar{\mathbf{W}}_{1}^\star, \bar{\mathbf{W}}_{2}^\star, \ldots, \bar{\mathbf{W}}_{n-1}^\star\), there exists a weight matrix \(\mathbf{W}_n\) such that the target provided in \eqref{eq:target_wn} can be approximated by the network whose output is given in \eqref{eq:mlfn_wn_out}. This is feasible if the number of neurons \(h_{n-1}\) is sufficiently large. We have
\begin{flalign}
&\boldsymbol{y}_{n} = \boldsymbol{y}_{\mathcal{NN}n}(\mathbf{W}_n,\bar{\boldsymbol{\phi}}_{n-1}^\star)
= \boldsymbol{\phi}_{n}({\mathbf{W}}_{n} \bar{\boldsymbol{\phi}}_{n-1}^\star)&\label{eq:mlfn_wn_y}
\end{flalign}
For learning of \(\mathbf{W}_{n}\), an incremental learning update law, referred to as \textit{one-layer update}, is developed to update the weights at that layer. 
In this algorithm, the weights in matrix \(\mathbf{W}_n\) are updated incrementally without inverting the activation functions in \(\boldsymbol{\phi}_{n}\) (and hence the update law is also called \textit{nonlinear}). In each step of training, we use one example of the training data. At the \(k^{\text{th}}\) step, \((\boldsymbol{z}(k), \boldsymbol{y}(k))\) are used, and the target for training \(\mathbf{W}_n\) in \eqref{eq:target_wn} is given as
\begin{flalign}
&\boldsymbol{y}_n(k) = \boldsymbol{y}(k) &\label{eq:target_wn_k}  
\end{flalign}
Equation \eqref{eq:mlfn_wn_y} can be rewritten as
\begin{flalign}
&\boldsymbol{y}_{n}(k) =  \boldsymbol{y}_{\mathcal{NN}n}(\mathbf{W}_n,\bar{\boldsymbol{\phi}}_{n-1}^\star(k))=\boldsymbol{\phi}_{n}({\mathbf{W}}_n \bar{\boldsymbol{\phi}}_{n-1}^\star(k))& \label{eq:mlfn_wn_yk}
\end{flalign} \vspace{-5mm}
\begin{flalign}
&\text{where } \bar{\boldsymbol{\phi}}_{n-1}^\star(k) =
\boldsymbol{\phi}_{n-1}(\bar{\mathbf{W}}_{n-1}^\star\boldsymbol{\phi}_{n-2}(& \nonumber\\
&{\;\;\;\;\;\;\;\;\;\;\;\;\;\;\;\;\;\;\;\;\;\;\;\;\;\;}\dots  \boldsymbol{\phi}_2(\bar{\mathbf{W}}_2^\star \boldsymbol{\phi}_1(\bar{\mathbf{W}}_1^\star \boldsymbol{z}(k)))\dots))& \label{eq:mlfn_phi_n_k} 
\end{flalign}
\noindent Let \(\hat{\mathbf{W}}_n(k)\) denote the estimated weight matrix at the \(k^{\text{th}}\) step of training, the estimated output \(\hat{\boldsymbol{y}}_{n}(k)\) at this \(k^{\text{th}}\) step is constructed as the direct output of this \({n}^\text{th}\) layer when its weight matrix is set at \(\hat{\mathbf{W}}_n(k)\):
\begin{eqnarray}
\hat{\boldsymbol{y}}_{n}(k) &=&\boldsymbol{y}_{\mathcal{NN}n}(\hat{\mathbf{W}}_n(k),\bar{\boldsymbol{\phi}}_{n-1}^\star(k))\nonumber\\
&=& \boldsymbol{\phi}_{n}(\hat{\mathbf{W}}_n(k) \bar{\boldsymbol{\phi}}_{n-1}^\star(k)) \label{eq:mlfn_wn_estyk} 
\end{eqnarray}
The output estimation error in learning of \(\mathbf{W}_n\) at the \(k^{\text{th}}\) step is defined as \(\boldsymbol{e}_{n}(k)=\boldsymbol{y}_{n}(k)-\hat{\boldsymbol{y}}_{n}(k)\).
Hence, from \eqref{eq:mlfn_wn_yk} and \eqref{eq:mlfn_wn_estyk} we have
\begin{flalign}
&\boldsymbol{e}_{n}(k) = \boldsymbol{y}_{\mathcal{NN}n}(\mathbf{W}_n,\bar{\boldsymbol{\phi}}_{n-1}^\star(k))&\nonumber\\
&{\;\;\;\;\;\;\;\;\;\;\;\;\;\;}-\boldsymbol{y}_{\mathcal{NN}n}(\hat{\mathbf{W}}_n(k),\bar{\boldsymbol{\phi}}_{n-1}^\star(k))  \label{eq:mlfn_general_er_wn}   \\
&{\;\;\;\;\;\;}={\boldsymbol{\phi}}_{n}(\mathbf{W}_n \bar{\boldsymbol{\phi}}_{n-1}^\star(k)) - {\boldsymbol{\phi}}_{n}(\hat{\mathbf{W}}_n(k) \bar{\boldsymbol{\phi}}_{n-1}^\star(k)) & \label{eq:mlfn_wn_er} \\
&\text{Let    }{\;\;} \boldsymbol{\delta}_{n}(k) \triangleq \mathbf{W}_n \bar{\boldsymbol{\phi}}_{n-1}^\star(k) - \hat{\mathbf{W}}_n(k) \bar{\boldsymbol{\phi}}_{n-1}^\star(k)\nonumber&\\
&{\;\;\;\;\;\;\;\;\;\;\;\;\;\;\;\;\;}=\Delta \mathbf{W}_n(k) \bar{\boldsymbol{\phi}}_{n-1}^\star(k)& \label{eq:mlfn_wn_deltak}
\end{flalign} 
where \(\Delta \mathbf{W}_n(k) = \mathbf{W}_n - \hat{\mathbf{W}}_n(k)\).  \vspace{-2mm}

\noindent Let us consider the relationship between \(\boldsymbol{e}_{n}(k)\) and \(\boldsymbol{\delta}_{n}(k)\):
If \({\boldsymbol{\phi}}_{n}\) is chosen as a vector of monotonically increasing activation functions whose derivatives are bounded by \(f_{\phi_{n}}\), {then:} \vspace{-3mm}
\begin{itemize}
	\item[i,]the corresponding elements of two vectors \( \boldsymbol{e}_{n}(k)\)  and \(\boldsymbol{\delta}_{n}(k)\)  have the same sign, i.e.
	\begin{equation}
	e_{n,i}(k)\delta_{n,i}(k) \geq 0,{\;} \forall i = 1..h_n \label{mlfn_wn_con_1}
	\end{equation}
	\item[ii,]the absolute value of an element of \(\boldsymbol{e}_{n}(k)\) is less than or equal to \(f_{\phi_{n}}\) times the corresponding element of
	\(\boldsymbol{\delta}_{n}(k)\), i.e.
	\begin{equation}
	|{e}_{n,i}(k)| \leq f_{\phi_{n}}|{\delta}_{n,i}(k)|, {\;} \forall i = 1..h_n  \label{mlfn_wn_con_2}
	\end{equation} 
\end{itemize}

The incremental learning law (one-layer update) to update the estimated weights based on the output estimation error is proposed as
\begin{equation}
\hat{\mathbf{W}}_n(k+1)=\hat{\mathbf{W}}_n(k)+ \mathbf{L}_{n}(k) \boldsymbol{e}_{n}(k)  \bar{\boldsymbol{\phi}}_{n-1}^{\star T}(k) \label{eq:mlfn_wn_updl} 
\end{equation}
where \(\mathbf{L}_{n}(k) \in \mathbb{R}^{h_{n} \times h_{n}}\) is a positive diagonal matrix; \(\bar{\boldsymbol{\phi}}_{n-1}^{\star}(k)\) is calculated using \eqref{eq:mlfn_phi_n_k}; \(\boldsymbol{e}_{n}(k)=\boldsymbol{y}_{n}(k)-\hat{\boldsymbol{y}}_{n}(k)\) with \(\boldsymbol{y}_n(k)\) and \(\hat{\boldsymbol{y}}_{n}(k)\) given in \eqref{eq:target_wn_k} and \eqref{eq:mlfn_wn_estyk}, respectively. {This update law \eqref{eq:mlfn_wn_updl} was developed so that the convergence could be ensured.}
Denoting \(\boldsymbol{w}_{n,i}\) as the \(i^{\text{th}}\) column vector of matrix \(\mathbf{W}_n\),  \(\hat{\boldsymbol{w}}_{n,i}(k)\) the  \(i^{\text{th}}\) column vector of \(\hat{\mathbf{W}}_n(k)\) and \(\bar{\phi}_{n-1,i}^\star(k)\) the \(i^{\text{th}}\) element of vector \(\bar{\boldsymbol{\phi}}_{n-1}^\star(k)\), the update law \eqref{eq:mlfn_wn_updl} can be rewritten in the vector form as
\begin{equation}
\hat{\boldsymbol{w}}_{n,i}(k+1)=\hat{\boldsymbol{w}}_{n,i}(k)+  \bar{\phi}_{n-1,i}^\star(k) \mathbf{L}_{n}(k)\boldsymbol{e}_{n}(k)  
\end{equation}

To show the convergence, we define an objective function as
\begin{equation}
V(k)= \sum_{i=1}^{h_{n-1}}{\Delta \boldsymbol{w}_{n,i}^T(k) \Delta \boldsymbol{w}_{n,i}(k)}  \label{eq:mlfn_wn_Vk}
\end{equation}
where \( \Delta \boldsymbol{w}_{n,i}(k) = \boldsymbol{w}_{n,i} -\hat{\boldsymbol{w}}_{n,i}(k)\). The objective function at the \((k+1)^{\text{th}}\) step is
\begin{align}
V(k+1) &= \sum_{i=1}^{h_{n-1}}{\Delta \boldsymbol{w}_{n,i}^T(k+1) \Delta \boldsymbol{w}_{n,i}(k+1)}  \nonumber
\end{align}
\vspace{-2mm}
\begin{align}
&= \sum_{i=1}^{h_{n-1}}(\Delta \boldsymbol{w}_{n,i}(k)- \bar{\phi}_{n-1,i}^\star(k)\mathbf{L}_{n}(k)\boldsymbol{e}_{n}(k) )^T\nonumber\\
&{\;\;\;\;\;\;\;\;\;\;\;\;}(\Delta \boldsymbol{w}_{n,i}(k)- \bar{\phi}_{n-1,i}^\star(k)\mathbf{L}_{n}(k)\boldsymbol{e}_{n}(k) ) 	
\end{align}
A change of the objective function value when the learning step goes from \(k^{\text{th}}\) to \((k+1)^{\text{th}}\)
\begin{eqnarray}
\Delta V(k) &=&V(k+1)-V(k) \nonumber\\
&=&\sum_{i=1}^{h_{n-1}} \Big(- \bar{\phi}_{n-1,i}^\star(k)\Delta\boldsymbol{w}_{n,i}^T(k) \mathbf{L}_{n}(k)\boldsymbol{e}_{n}(k)  \nonumber\\
&&{\;\;-\;} \bar{\phi}_{n-1,i}^\star(k)\boldsymbol{e}_{n}^T(k)\mathbf{L}_{n}^T(k)\Delta\boldsymbol{w}_{n,i}(k) \nonumber\\
&&{\;\;+\;}  \bar{\phi}_{n-1,i}^{\star 2}(k)\boldsymbol{e}_{n}^T(k)\mathbf{L}_{n}^T(k)\mathbf{L}_{n}(k) \boldsymbol{e}_{n}(k) \Big)   \label{eq:mlfn_wn_deltaV} 
\end{eqnarray}
From \eqref{eq:mlfn_wn_deltak}, we have \(\boldsymbol{\delta}_{n}(k) =\Delta \mathbf{W}_{n}(k) \bar{\boldsymbol{\phi}}_{n-1}^\star(k)\\=\sum_{i=1}^{h_{n-1}} {\Delta\boldsymbol{w}_{n,i}(k)\bar{\phi}_{n-1,i}^\star(k)} \), hence
\begin{align}
\Delta V(k)={\;} &{-\;} \boldsymbol{\delta}_{n}^T(k) \mathbf{L}_{n}(k)\boldsymbol{e}_{n}(k)  - \boldsymbol{e}_{n}^T(k)\mathbf{L}_{n}^T(k)\boldsymbol{\delta}_{n}(k) \nonumber\\
&{+\;} \mu_{n-1}(k)\boldsymbol{e}_{n}^T(k)\mathbf{L}_{n}^T(k)\mathbf{L}_{n}(k) \boldsymbol{e}_{n}(k)   
\end{align}
{with \(\bar{\mu}_{n-1}^{\star}(k)\triangleq\sum_{i=1}^{h_{n-1}} \bar{\phi}_{n-1,i}^{\star 2}(k)\).}
From the properties stated in \eqref{mlfn_wn_con_1}, \eqref{mlfn_wn_con_2}, we have the following inequality since \(\mathbf{L}_{n}(k)\) is a positive diagonal matrix
\begin{equation}
\boldsymbol{\delta}_{n}^T(k) \mathbf{L}_{n}(k)\boldsymbol{e}_{n}(k) \geq \frac{1}{f_{\phi_{n}}}\boldsymbol{e}_{n}^T(k) \mathbf{L}_{n}(k)\boldsymbol{e}_{n}(k) 
\end{equation}
which finally gives
\begin{flalign}
\Delta V(k) &\leq  -{\;}\boldsymbol{e}_{n}^T(k)\Big(\frac{2}{f_{\phi_{n}}}\mathbf{L}_{n}(k)&\nonumber\\
&{\;\;\;\;\;\;\;\;\;\;\;\;\;\;\;\;\;}-\bar{\mu}_{n-1}^{\star}(k)\mathbf{L}_{n}^T(k)\mathbf{L}_{n}(k)\Big)\boldsymbol{e}_{n}(k)& \label{eq:mlfn_wn_deltaV_final} 
\end{flalign}
When \(\mathbf{L}_{n}(k)\) is chosen such that
\begin{equation}
\frac{2}{f_{\phi_{n}}}\mathbf{L}_{n}(k)-\bar{\mu}_{n-1}^{\star}(k)\mathbf{L}_{n}^T(k)\mathbf{L}_{n}(k) >0 \label{eq:mlfn_wn_con_L}
\end{equation} 
{then \( \Delta V(k) \) is non-positive for any \(\boldsymbol{e}_{n}(k)\). That means, the value of {the} objective function is a non-increasing sequence \(V(k+1)\leq V(k)\). Moreover, since the function \(V(k)\) is non-negative, which means {that} it is bounded from below, we have \( \Delta V(k)\) converges to 0 as \(k\) tends to infinity. Thus, from \eqref{eq:mlfn_wn_deltaV_final}, \( \boldsymbol{e}_{n}(k)\) converges to 0 as \(k \) tends to infinity.} \vspace*{-2mm}

{\color{black}\textbf{Remark 2}: Updating one layer of weights is a common technique in neural network based control \cite{sanner1992gaussian,yang2014neural,li2014adaptive,he2016adaptive,yang2016neural,licitra2019single,qiu2019command,mu2019learning,yang2019force,lyu2020data}.  In addition, for single-hidden layer networks (i.e., two-layer networks) with fixed weights in input-to-hidden layer, it has been shown that an RBF network is a universal approximator \cite{park1991universal}, and a Fourier network {can approximate any square integrable function} to any desired accuracy \cite{gallant1988there}. For networks with random weights or features in the input-to-hidden layer, with sufficiently large number of neurons in the hidden layer, updating only the output weights can also make the networks universal approximators \cite{chong2019closer,rahimi2008weighted,rahimi2008uniform,huang2006universal}. In the case of insufficient number of hidden nodes, the convergence can also be analyzed (see Remark 4).} \vspace*{-2mm}

{\textbf{Remark 3}: With a symmetric positive-definite matrix \(\mathbf{L}_{n}(k)\), the condition \eqref{eq:mlfn_wn_con_L} is met if
	\begin{equation}
	\frac{2}{f_{\phi_{n}}}\mathbf{I}-\bar{\mu}_{n-1}^{\star}(k)\mathbf{L}_{n}(k) > 0 \label{eq:mlfn_wn_con_L_pd} 
	\end{equation} 
where \(\mathbf{I} \in \mathbb{R}^{h_{n} \times h_{n}}\) is the identity matrix. As \(\mathbf{L}_{n}(k)\) is also diagonal, condition \eqref{eq:mlfn_wn_con_L} or \eqref{eq:mlfn_wn_con_L_pd} suggests that the matrix \(\mathbf{L}_{n}(k)\) can be adjusted to smaller values so that the left hand side (LHS) of either inequality is positive-definite. Therefore, one possibility is that \(\mathbf{L}_{n}(k)\) is initialized to an arbitrary value and the LHS of \eqref{eq:mlfn_wn_con_L} or \eqref{eq:mlfn_wn_con_L_pd} is checked for positive definiteness in each learning step. If the condition is not met, \(\mathbf{L}_{n}(k)\) is then reduced so that the condition is met. Another way for finding the positive diagonal matrix \(\mathbf{L}_{n}(k)\) is by looking at its diagonal entries \({l}_{n\iota}(k)\), \(\iota=1..h_n\). Noting that the LHS of  \eqref{eq:mlfn_wn_con_L} or \eqref{eq:mlfn_wn_con_L_pd} is also diagonal. Hence, {\eqref{eq:mlfn_wn_con_L} and \eqref{eq:mlfn_wn_con_L_pd}} can be ensured if
\begin{equation}
	\frac{2}{f_{\phi_{n}}}>\bar{\mu}_{n-1}^{\star}(k){l}_{n\iota}(k), {\;} \forall \iota=1..h_n \label{eq:mlfn_wn_con_L_} 
\end{equation}
The condition given by \eqref{eq:mlfn_wn_con_L_} can be easily implemented to find the diagonal entries and thus the matrix \(\mathbf{L}_{n}(k)\) can be established.
Also noting that the range of the choice of \({l}_{n\iota}(k)\) can be made larger with smaller \(f_{\phi_{n}}\) or \(\bar{\mu}_{n-1}^{\star}(k)\). The former can be achieved by changing the output activation functions and the latter can be done by normalizing the inputs of previous layers, or setting the ranges of the weights of previous layers to smaller values, or making the hidden activation functions bounded with smaller values.}  \vspace*{-2mm}

{\textbf{Remark 4}: In case of having an insufficient number of hidden neurons and/or in presence of measurement noise, by extending the time domain results in \cite{kawamura1988local}, it can be shown that if
	\begin{align} 
	&\Vert {\boldsymbol{e}}_{n}(k) \Vert  \geq  \frac{b}{2\left( \frac{2L_{m}}{f_{\phi_{n}M}}-c_M L_{M}^2\right)} \Bigg[\frac{4L_{m}}{f_{\phi_{n}M}}+\frac{2L_{M}}{f_{\phi_{n}m}}+\nonumber\\
	&\medmath{+\sqrt{\frac{8L_{m}}{f_{\phi_{n}M}}c_M L_{M}^2+ \frac{8L_{M}}{f_{\phi_{n}m}}\left( \frac{4L_{m}}{f_{\phi_{n}M}}-c_M L_{M}^2\right)+\left(\frac{2L_{M}}{f_{\phi_{n}m}}\right)^2}}\Bigg]   \label{eq:bound}
	\end{align}\vspace{-3mm}
	\begin{flalign}
	&\text{and\;\;\;\;\;\;}\frac{2L_{m}}{f_{\phi_{n}M}}-c_M L_{M}^2 > 0&   
	\end{flalign}
	then \(\Delta V(k) \leq 0\). 
	Here \(b\) is the upper bound of the NN approximation error and measurement noise, \(L_{m}\) and \(L_{M}\) are respectively the minimum and maximum eigenvalues of the matrix \(\mathbf{L}_{n}(k)\) for \(\forall k\), \(f_{\phi_{n}m}\) and \(f_{\phi_{n}M}\) are the lower and the upper bounds of the derivative of \({\boldsymbol{\phi}}_{n}\) respectively, and \(c_M\) is the maximum value of all \(\bar{\mu}_{n-1}^{\star}(k)\). Therefore, there exists an ultimate bound such that the error always stay within the bound after reaching it. Noting {from \eqref{eq:bound}} that the bound would tend to zero when the NN approximation error and noise tend to zero.} 
\vspace{-2mm}

\subsubsection{Learning of the hidden weights \(\mathbf{W}_{j}\) with \(n-1 \geq j \geq 2\)}
\begin{figure}[ht]
	\centering
	\includegraphics[width=2.8in]{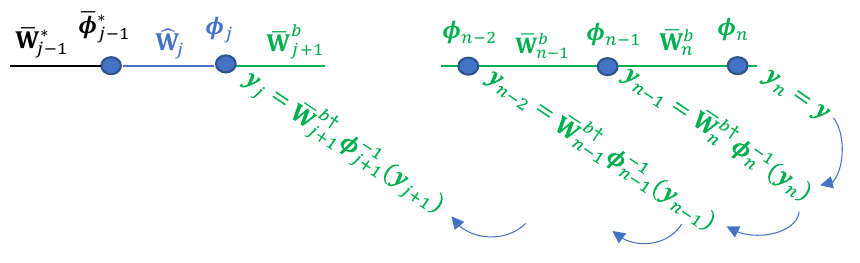}
	\caption{Backward transmission of the target in inverse layer-wise learning, from the output layer to the inner layers.}
	\label{fig:back_trans}  \vspace{-3mm}
\end{figure}

After \(\mathbf{W}_{n}\)  has been learned and its value in this backward stage has been obtained as \(\bar{\mathbf{W}}_{n}^{b}\), the target for the \((n-1)^\text{th}\) layer is calculated based on the left hand side of \eqref{eq:learn_wn-1} as \( \boldsymbol{y}_{n-1} = \bar{\mathbf{W}}_{n}^{b\dagger}\boldsymbol{\phi}_{n}^{-1}(\boldsymbol{y}_{n})\). Generally, the target for learning of \(\mathbf{W}_{j}\) (or the \(j^\text{th}\) layer), denoted as \(\boldsymbol{y}_{j}\), can be achieved by calculating backwardly from the target for the last layer \(\boldsymbol{y}_n = \boldsymbol{y}\), using the left-hand sides of equations from \eqref{eq:learn_wn} to \eqref{eq:learn_w0}. We have
\begin{equation}
\boldsymbol{y}_{j} = \bar{\mathbf{W}}_{j+1}^{b\dagger}\boldsymbol{\phi}_{j+1}^{-1}(\boldsymbol{y}_{j+1})  \text{ with }  n-1 \geq j \geq 1 \label{eq:target_wj}
\end{equation}
A graphical illustration of the backward calculation of the target is shown in Fig. \ref{fig:back_trans}. \vspace{-2mm}

\noindent The input of the \(j^\text{th}\) layer is computed similarly to \eqref{eq:input_wn} using \eqref{eq:mlfn_output_j-1}
\begin{flalign}
\bar{\boldsymbol{\phi}}_{j-1}^\star \triangleq \boldsymbol{\phi}_{j-1}(\bar{\mathbf{W}}_{j-1}^\star\boldsymbol{\phi}_{j-2}(\dots  \boldsymbol{\phi}_2(\bar{\mathbf{W}}_2^\star \boldsymbol{\phi}_1(\bar{\mathbf{W}}_1^\star \boldsymbol{z}))\dots)) \label{eq:input_wj}
\end{flalign}
Hence, the output of this \(j^\text{th}\) layer is given as
\begin{eqnarray}
\boldsymbol{y}_{\mathcal{NN}j}(\mathbf{W}_j,\bar{\boldsymbol{\phi}}_{j-1}^\star)
= \boldsymbol{\phi}_{j}({\mathbf{W}}_{j} \bar{\boldsymbol{\phi}}_{j-1}^\star)\label{eq:mlfn_wj_out}
\end{eqnarray}
There exists a weight matrix \(\mathbf{W}_j\) such that the target provided in \eqref{eq:target_wj} can be approximated by the network whose output is given in \eqref{eq:mlfn_wj_out}
\begin{eqnarray}
\boldsymbol{y}_{j} = \boldsymbol{y}_{\mathcal{NN}j}(\mathbf{W}_j,\bar{\boldsymbol{\phi}}_{j-1}^\star)
= \boldsymbol{\phi}_{j}({\mathbf{W}}_{j} \bar{\boldsymbol{\phi}}_{j-1}^\star)\label{eq:mlfn_wj_y}
\end{eqnarray}
Similar to learning of \(\mathbf{W}_{n}\),  the weights in matrix \(\mathbf{W}_j\) are updated incrementally without inverting the activation functions \(\boldsymbol{\phi}_{j}\). The functions \({\boldsymbol{\phi}}_{j}\) are chosen to be monotonically increasing activation functions and their derivatives are bounded by \(f_{\phi_{j}}\). At the \(k^{\text{th}}\) step of training, equation \eqref{eq:target_wj} can be rewritten as
\begin{equation}
\boldsymbol{y}_{j}(k) = \bar{\mathbf{W}}_{j+1}^{b\dagger}\boldsymbol{\phi}_{j+1}^{-1}(\boldsymbol{y}_{j+1}(k))   \label{eq:target_wj_k}
\end{equation}
and equation \eqref{eq:mlfn_wj_y} can be rewritten as
\begin{equation}
\boldsymbol{y}_{j}(k) =  \boldsymbol{y}_{\mathcal{NN}j}(\mathbf{W}_j,\bar{\boldsymbol{\phi}}_{j-1}^\star(k))=\boldsymbol{\phi}_{j}({\mathbf{W}}_j \bar{\boldsymbol{\phi}}_{j-1}^\star(k)) \label{eq:mlfn_wj_yk}
\end{equation} \vspace{-5mm}
\begin{flalign}
&\text{where } \bar{\boldsymbol{\phi}}_{j-1}^\star(k)= \boldsymbol{\phi}_{j-1}(\bar{\mathbf{W}}_{j-1}^\star\boldsymbol{\phi}_{j-2}(&\nonumber\\
&{\;\;\;\;\;\;\;\;\;\;\;\;\;\;\;\;\;\;\;\;\;\;\;\;\;\;}\dots  \boldsymbol{\phi}_2(\bar{\mathbf{W}}_2^\star \boldsymbol{\phi}_1(\bar{\mathbf{W}}_1^\star \boldsymbol{z}(k)))\dots))& \label{eq:mlfn_phi_j_k} 
\end{flalign}
\noindent Let \(\hat{\mathbf{W}}_j(k)\) denote the estimated weight matrix at the \(k^{\text{th}}\) step of training, the estimated output \(\hat{\boldsymbol{y}}_{j}(k)\) at the \(k^{\text{th}}\) step is constructed as the direct output of this \({j}^\text{th}\) layer when its weight matrix is set at \(\hat{\mathbf{W}}_j(k)\):
\begin{equation}
\hat{\boldsymbol{y}}_{j}(k) =\boldsymbol{y}_{\mathcal{NN}j}(\hat{\mathbf{W}}_j(k),\bar{\boldsymbol{\phi}}_{j-1}^\star(k))= \boldsymbol{\phi}_{j}(\hat{\mathbf{W}}_j(k) \bar{\boldsymbol{\phi}}_{j-1}^\star(k)) \label{eq:mlfn_wj_estyk}
\end{equation}
The output estimation error in learning of \(\mathbf{W}_j\) at the \(k^{\text{th}}\) step is defined as \(\boldsymbol{e}_{j}(k)=\boldsymbol{y}_{j}(k)-\hat{\boldsymbol{y}}_{j}(k)\). Hence,
\begin{equation}
\boldsymbol{e}_{j}(k)= \boldsymbol{y}_{\mathcal{NN}j}(\mathbf{W}_j,\bar{\boldsymbol{\phi}}_{j-1}^\star(k))-\boldsymbol{y}_{\mathcal{NN}j}(\hat{\mathbf{W}}_j(k),\bar{\boldsymbol{\phi}}_{j-1}^\star(k)) \label{eq:mlfn_general_er}
\end{equation}

The incremental learning law to update the estimated weights based on the output estimation error is proposed as \vspace{-2mm}
\begin{equation}
\hat{\mathbf{W}}_j(k+1)=\hat{\mathbf{W}}_j(k)+ \mathbf{L}_{j}(k) \boldsymbol{e}_{j}(k)  \bar{\boldsymbol{\phi}}_{j-1}^{\star T}(k) \label{eq:mlfn_wj_updl} 
\end{equation}
where \(\mathbf{L}_{j}(k) \in \mathbb{R}^{h_{j} \times h_{j}}\) is a positive diagonal matrix that satisfies the following condition
\begin{align}
\frac{2}{f_{\phi_{j}}}\mathbf{L}_{j}(k)-\bar{\mu}_{j-1}^{\star}(k)\mathbf{L}_{j}^T(k)\mathbf{L}_{j}(k) >0 \label{eq:mlfn_wj_con_L}
\end{align}
with \(\bar{\mu}_{j-1}^{\star}(k)\triangleq\sum_{i=1}^{h_{j-1}}{ \bar{\phi}_{j-1,i}^{\star 2}(k)}\); \(\bar{\phi}_{j-1,i}^{\star 2}(k)\) being the \(i^\text{th}\) element of vector \(\bar{\boldsymbol{\phi}}_{j-1}^{\star}(k)\); \(\bar{\boldsymbol{\phi}}_{j-1}^{\star}(k)\) is calculated using \eqref{eq:mlfn_phi_j_k}; \(\boldsymbol{e}_{j}(k)=\boldsymbol{y}_{j}(k)-\hat{\boldsymbol{y}}_{j}(k)\) with \(\boldsymbol{y}_j(k)\) and \(\hat{\boldsymbol{y}}_{j}(k)\) given in \eqref{eq:target_wj_k} and \eqref{eq:mlfn_wj_estyk}, respectively. 
{Similar to \eqref{eq:mlfn_wn_con_L_}, the condition can be rewritten as (see Remark 3)
	\begin{equation}
	\frac{2}{f_{\phi_{j}}} > \bar{\mu}_{j-1}^{\star}(k){l}_{j\iota}(k) , {\;} \forall \iota=1..h_{j} \label{eq:mlfn_wj_con_L_explicit} 
	\end{equation}}
Similarly to the case of \(\mathbf{W}_n\), it can be shown that \( \boldsymbol{e}_{j}(k)\) converges as \(k \) increases.\vspace{-2mm}

\subsubsection{Learning of the input weights \(\mathbf{W}_{1}\)} \vspace{-2mm}

Learning of \(\mathbf{W}_{1}\) is done in the same way as learning of \(\mathbf{W}_{j}\) above by setting \(j = 1\). One point to take note is that the denotations of \(\bar{\boldsymbol{\phi}}^\star_{j-1} \) and \(h_{j-1}\) in above demonstrations would become \(\bar{\boldsymbol{\phi}}^\star_{0} \) and \(h_{0}\) which do not exist. However, it is possible to consider that \(\bar{\boldsymbol{\phi}}^\star_{0} \triangleq \boldsymbol{z}\) (\(\bar{\phi}^\star_{0,i} \triangleq {z}_i\)) and \(h_{0} \triangleq m\) (\(m\) is the dimension of the input vector \(\boldsymbol{z}\)). After learning, we get \(\bar{\mathbf{W}}_{1}^b\). The backward stage stops.  \vspace{-2mm}

\subsection{Forward Stage of Inverse Layer-wise Learning} \label{sec:forward} \vspace{-2mm}

In the forward stage, the network is trained forwardly from the input layer to the output layer. The inverse layer-wise learning goes forwards from \(\mathbf{W}_2\) until the output weights \(\mathbf{W}_n\) are retrained. \vspace*{-2mm}

In this stage, the learning can be conducted similarly to the backward stage for every layer from \(\mathbf{W}_{2}\) to \(\mathbf{W}_{n}\). The only difference is that in the backward stage, the input of the \(j^\text{th}\) layer was defined based on the initialized values of \(\bar{\mathbf{W}}_1^{\star}, \bar{\mathbf{W}}_2^{\star}, \ldots, \bar{\mathbf{W}}_{j-1}^{\star}\) as in \eqref{eq:input_wj}, but it is now defined based on the values of weights \(\bar{\mathbf{W}}_1, \bar{\mathbf{W}}_2, \ldots, \bar{\mathbf{W}}_{j-1}\) that have been relearned before \({\mathbf{W}}_{j}\)
\begin{flalign}
\bar{\boldsymbol{\phi}}_{j-1} = \boldsymbol{\phi}_{j-1}(\bar{\mathbf{W}}_{j-1}\boldsymbol{\phi}_{j-2}(\dots \boldsymbol{\phi}_2&(\bar{\mathbf{W}}_2 \boldsymbol{\phi}_1(\bar{\mathbf{W}}_1 \boldsymbol{z}))\dots)) \nonumber\\
&\text{ with } 2\leq j \leq n \label{eq:input_wj_forward}
\end{flalign}
with noting that \(\bar{\mathbf{W}}_{1} = \bar{\mathbf{W}}_{1}^b\) since there is no forward learning for \(\mathbf{W}_{1}\). \vspace{-2mm}

\subsection{The Step-by-Step Algorithm} \vspace{-2mm}
This part summarizes the inverse layer-wise learning of MLFNs. In each layer, the weights are updated incrementally by a one-layer update law. 
The details of the inverse layer-wise learning algorithm are as follows \vspace{-2mm}
\begin{enumerate}
	\item[(i)] Initialization: Randomly assign \(\mathbf{W}_1, \mathbf{W}_2, \dots, \mathbf{W}_{n}\).
	\item [(ii)] Train \(\mathbf{W}_{n}\) using update law \eqref{eq:mlfn_wn_updl}  \(\to\) obtain \(\bar{\mathbf{W}}_{n}^b\)
	\item [(iii)] Backward looping: Learning of \({\mathbf{W}}_{j}\) (from \({\mathbf{W}}_{n-1}\) to \({\mathbf{W}}_{1}\))
	\begin{itemize}
	\item [(a)] Set \(j=n-1\)
	\item [(b)] Train \(\mathbf{W}_{j}\) using update law \eqref{eq:mlfn_wj_updl} \(\to\) obtain \(\bar{\mathbf{W}}_{j}^b\)
	\item [(c)] Decrease \(j\) by 1 and go to step (b) if \(j \geq 1\)
	\end{itemize}
	\item [(iv)] Forward looping: Learning of \({\mathbf{W}}_{j}\) again (from \({\mathbf{W}}_{2}\) to \({\mathbf{W}}_{n}\))
	\begin{itemize}
	\item [(d)] Set \(j=2\)
	\item [(e)] Train \(\mathbf{W}_{j}\) using update law  \eqref{eq:mlfn_wj_updl} with noting that \(\bar{\boldsymbol{\phi}}_{j-1}\) in \eqref{eq:input_wj_forward} is used instead of \(\bar{\boldsymbol{\phi}}^\star_{j-1}\) \(\to\) obtain \(\bar{\mathbf{W}}_{j}\)
	\item [(f)] Increase \(j\) by 1 and go to step (e) if \(j \leq n\)
	\end{itemize}
\vspace{-2mm}
\end{enumerate}

{In learning of each layer, each time when all examples in the training data have been used is called a loop. In practice, it can take many loops to train a layer, which means that each training example can be called upon many times.} \vspace*{-2mm}

\subsection{Classification on MNIST Database and Performance Issue of Inverse Layer-wise Learning}  \label{sec: mnist_Ex1} \vspace{-2mm}
MNIST database {\cite{lecun1998mnist}} was used for assessing the performance of the inverse layer-wise learning algorithm to understand the issues associated with it. \vspace{-2mm}
{\begin{itemize}
	\item MNIST is a classical database of handwritten digits. It contains 60,000 images in the training set and 10,000 images in the test set. All of the black and white images are centered and have the same size of 28\(\times\)28 pixels.
	\vspace{-3mm}
\end{itemize} }
\subsubsection{Network structure and learning results} \vspace{-2mm}
{In this classification task with MNIST database, we used the network with the following properties:} \vspace{-3mm}
\begin{itemize}
	\item A 3-hidden layer network with structure 784-300-100-50-10 (300 units, 100 units, and 50 units in hidden layers). The activation functions at hidden layers are modified softplus \(f(x)=log(0.8+e^x)\) as suggested in \cite{toh2018analytic}, which has its inverse function as \(log(e^x-0.8)\). The activation functions at output are sigmoid  \(f(x)={1}/{(1+e^{-x})}\).
\vspace{-3mm}
\end{itemize} 
In this network, there are 4 weight matrices to be learned: \(\mathbf{W}_{1}, \mathbf{W}_{2}\), \(\mathbf{W}_{3}\) and \(\mathbf{W}_{4}\). Before training, \(\mathbf{W}_{1} \text{ to } \mathbf{W}_{4}\) were randomly initialized. After that, learning process began with learning of \(\mathbf{W}_{4} \to\) learning of \(\mathbf{W}_{3} \to\) learning of \(\mathbf{W}_{2} \to\) learning of \(\mathbf{W}_{1} \to\) relearning of \(\mathbf{W}_{2} \to\) relearning of \(\mathbf{W}_{3} \to\)  relearning of \(\mathbf{W}_{4}\). In learning of each layer, the gain matrix \(\mathbf{L}\) was not a predefined matrix.  Instead, it was calculated from the respective conditions in \eqref{eq:mlfn_wn_con_L} and \eqref{eq:mlfn_wj_con_L} to ensure the convergence of the learning process {by using \eqref{eq:mlfn_wn_con_L_} and \eqref{eq:mlfn_wj_con_L_explicit}.} The number of loops for training of each layer was 20.
\begin{table}[ht]
	\normalsize
	\begin{center}
		\caption{Training \& test accuracies (\%) by inverse layer-wise learning vs. SGD}
		\label{tbl: mnist_1}
		\setlength\extrarowheight{-2.5pt}
		\begin{tabular}{l|c|r}
			\textbf{ }&\textbf{Training} & \textbf{Test}\\
			\hline			
			Inverse layer-wise learning & 92.37 & 92.04\\
			Stochastic gradient descent &99.80 & 98.36\\	
			\hline 
		\end{tabular}
	\end{center}
\vspace{-2mm}
\end{table}

The results in Table \ref{tbl: mnist_1} show that although convergence can be achieved, the accuracy after training by inverse layer-wise learning algorithm is not desirable as compared to the stochastic gradient descent (SGD) method ( 92.04\% compared to 98.36\% on the test set). Noting that for the purpose of comparing with the best performance, the results for SGD method in this paper were achieved by observing the accuracy directly on the test set while training and no validation set was used.\vspace{-2mm}

\subsubsection{Discussions on the results of inverse layer-wise learning} \vspace{-2mm}

The inverse layer-wise learning method is non-error-based, which means that the error at the output layer of the MLFN is not directly used to adjust all layers of weights (the output error is only used directly for training the last layer). Instead, the target is transmitted backwards to previous layers in a backward transmission process (please refer to \eqref{eq:target_wj} and Fig. \ref{fig:back_trans}). 
Though convergence can now be ensured in leaning, this backward transmission of target causes some possible problems that lead to a trade off in accuracy as compared to backpropagation (like SGD) method. One of the reasons can stem from the fact that the modified softplus and sigmoid are only invertible within their ranges. This causes the distortions in the target values when transmitted backwardly. Even when all the nonlinear functions are fully invertible and the targets for previous layer are right values, training only a layer of weights may not help in fitting the target for that layer. Therefore the inverse layer-wise learning method can be used in cases where a trade off in performance is acceptable while ensuring the convergence of the learning systems. In the next section, we present a learning method to achieve a good balance between accuracy and convergence. \vspace{-2mm}

\section{Forward Progressive Learning of Multilayer Feedforward Neural Networks} \label{sec: main_2} \vspace{-2mm}
In this section, we develop a forward progressive leaning (FPL) method based on the layer-wise methodology referred in \cite{bengio2007greedy,erhan2010does,hettinger2017forward}. Unlike those works \cite{bengio2007greedy,erhan2010does,hettinger2017forward} where there is no convergence analysis, the convergence can be analyzed in our proposed FPL method. Our main aim here is to develop an output-error-based layer-wise learning algorithm so as to overcome the drawback of the inverse layer-wise leaning method in section \ref{sec: main_1}. \vspace{-2mm}
\begin{figure}[t]
	\centering
	\includegraphics[width=2.7in]{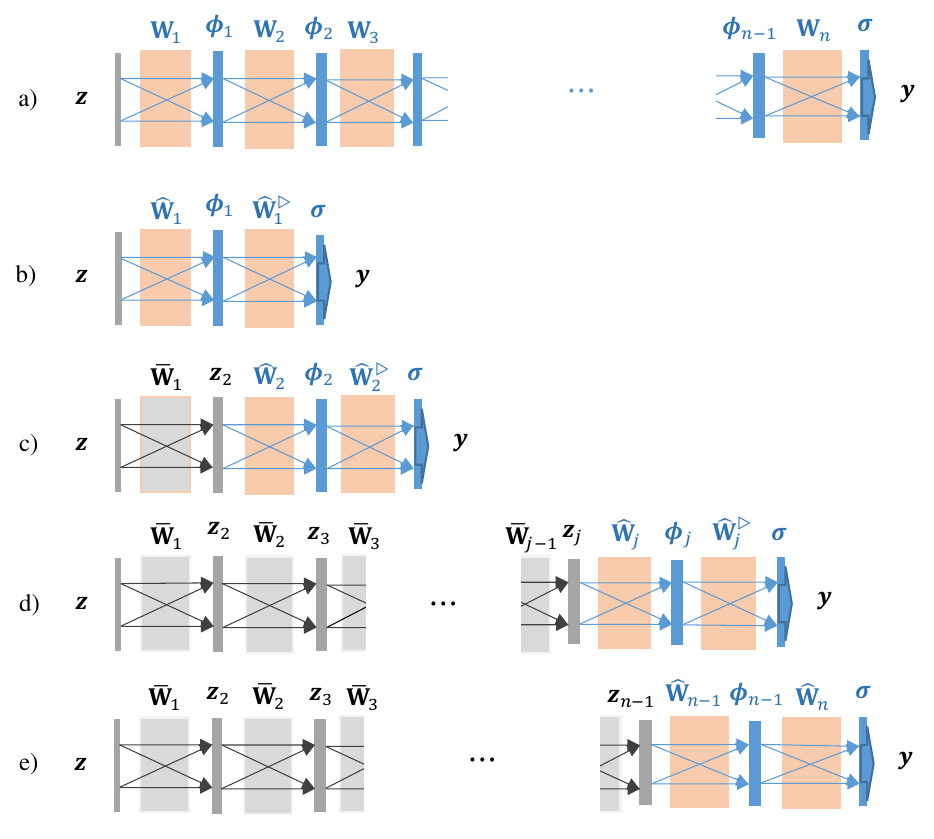} 
	\caption{Forward progressive learning (FPL) method, where an $n$-layer network is trained layer-wisely through learning of $(n-1)$ two-layer networks. Each two-layer network is trained in 2 phases: pre-training (subsection \ref{sec:pre-train}) and fine-tuning (subsection \ref{sec:fine_tuning}) so as to guarantee the convergence.}
	\label{fig:FT}
\end{figure}

Fig. \ref{fig:FT} illustrates the processing details of the algorithms. The overall structure of the deep network with \(n\) weight matrices to be learned is shown in Fig \ref{fig:FT}(a). The FPL starts with learning of the weights in the first layer \({\mathbf{W}}_{1}\) based on an SLFN (or two-layer network) where the first hidden layer is directly connected to the output, as shown in fig \ref{fig:FT}(b). Two weight matrices, an input weight matrix \({\mathbf{W}}_{1}\) and a pseudo output weight matrix \(\mathbf{W}_1^{\vartriangleright}\), are learned simultaneously by a two-layer algorithm. After learning, the matrix \(\bar{\mathbf{W}}_{1}\) is kept to form new input for the next layer while the pseudo output weight matrix \(\bar{\mathbf{W}}_1^{\vartriangleright}\) is discarded. Fig \ref{fig:FT}(c) shows the learning of the second layer \({\mathbf{W}}_{2}\) based on a second SLFN with two weight matrices \({\mathbf{W}}_{2}\) and \({\mathbf{W}}_{2}^{\vartriangleright}\). New input \(\boldsymbol{z}_2\) of this SLFN has been formed by passing \(\boldsymbol{z}\) through fixed \(\bar{\mathbf{W}}_{1}\). Similarly, after training we keep \(\bar{\mathbf{W}}_{2}\)
and discard \(\bar{\mathbf{W}}_{2}^{\vartriangleright}\). Fig \ref{fig:FT}(d) shows the learning of the \(j^\text{th}\) layer \({\mathbf{W}}_{j}\) with input \(\boldsymbol{z}_j\) and target output \(\boldsymbol{y}\). The FPL continues until the learning of \({\mathbf{W}}_{n-1}\) takes place as shown in Fig \ref{fig:FT}(e), where the pseudo output weights are no longer needed. Instead, the true output weights of the deep net  \({\mathbf{W}}_{n}\) is used and trained together with \({\mathbf{W}}_{n-1}\). After training this SLFN, the FPL ends. \vspace*{-3mm}

\begin{figure}[t]
	\centering
	\includegraphics[width=2.5in]{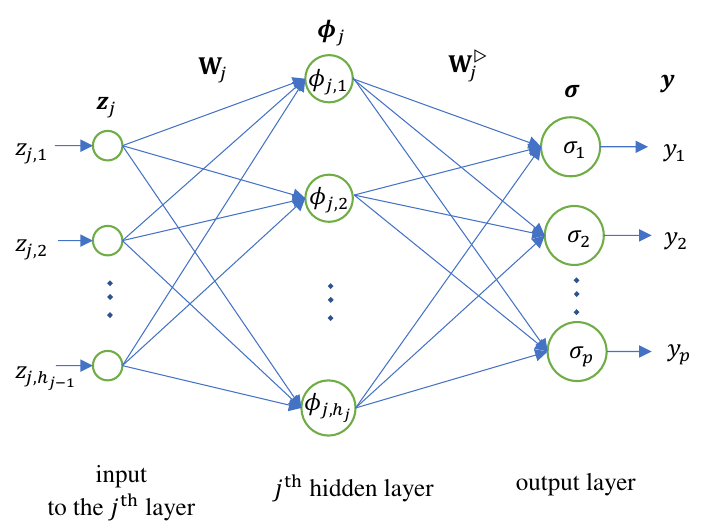}
	\caption{The single hidden layer feedforward network (or two-layer network) for new added \(j^\text{th}\) hidden layer in forward progressive learning. The input $\boldsymbol{z}_j$ of this network is calculated by forwardly propagating $\boldsymbol{z}$ through previous layers which have been learned, while its target is  the overall target $\boldsymbol{y}$ directly.}
	\label{fig:rbfnn}
	\vspace{-2mm}
\end{figure}

We now consider a general case when the \(j^\text{th}\) hidden layer is added. The structure of the SLFN in this case is shown in Fig. \ref{fig:rbfnn}. Since \(\bar{\mathbf{W}}_{1}, \ldots, \bar{\mathbf{W}}_{j-1}\) have been trained, the input to this SLFN can be calculated similarly to \eqref{eq:input_wj}
\begin{align}
\boldsymbol{z}_j &\triangleq \bar{\boldsymbol{\phi}}_{j-1} = \boldsymbol{\phi}_{j-1}(\bar{\mathbf{W}}_{j-1}\boldsymbol{\phi}_{j-2}( \nonumber\\
&{\;\;\;\;\;\;\;\;\;\;\;\;\;\;\;\;\;\;\;\;\;\;\;\;}\dots  \boldsymbol{\phi}_2(\bar{\mathbf{W}}_2 \boldsymbol{\phi}_1(\bar{\mathbf{W}}_1 \boldsymbol{z}))\dots)) \label{eq:input_wj_FPL} 
\end{align}
\noindent The output of the network as shown in Fig. \ref{fig:rbfnn} can be expressed as follows:
\begin{equation}
\boldsymbol{y}_{\mathcal{NN}}(\mathbf{W}_j,\mathbf{W}_j^{\vartriangleright},\boldsymbol{z}_j) = \boldsymbol{\sigma}(\mathbf{W}_j^{\vartriangleright} \boldsymbol{\phi}_{j}(\mathbf{W}_j \boldsymbol{z}_j)) \label{eq:rbf_output}
\end{equation}
Each SLFN is trained in two phases. In the first phase, called pre-training, the proposed one-layer update law in section \ref{sec: main_1} is adopted to pre-train the SLFN. In the second phase, called fine-tuning, a two-layer update law is developed to fine-tune the weights of the SLFN. \vspace{-2mm}
\subsection{Pre-training of Single Hidden Layer Feedforward Networks} \label{sec:pre-train} \vspace{-2mm}
The purpose of the pre-training phase is to achieve a sufficiently small estimation error, before further improvement is achieved by a fine-tunning phase. \vspace{-3mm}

The proposed one-layer update law for training the last layer of an MLFN is used to pre-train the SLFN. The steps of training are as follows:\vspace{-3mm}
\begin{itemize}
	\item Initialization: Randomly assign \(\mathbf{W}_j\) and \(\mathbf{W}_j^{\triangleright}\) to \(\bar{\mathbf{W}}_j^{\star}\) and  \(\bar{\mathbf{W}}_j^{\triangleright \star}\) respectively.
	\item \(\mathbf{W}_j^{\triangleright}\) is trained using update law \eqref{eq:mlfn_wn_updl}  to obtain \(\bar{\mathbf{W}}_j^{\triangleright}\).
	\vspace*{-2mm}
\end{itemize} 

After \(\mathbf{W}_j^{\triangleright}\) has been pre-trained, the entire SLFN will be trained one more time in a fine-tuning phase. \vspace*{-2mm}

Noting that the inverse layer-wise algorithm presented in the previous section can also be used to achieved this aim. \vspace*{-2mm}

\subsection{Fine-Tuning of Single Hidden Layer Feedforward Networks} \label{sec:fine_tuning} \vspace{-2mm}

In this subsection, a \textit{two-layer update} law is developed to update concurrently both the input weights \(\mathbf{W}_j\) and the output weights \(\mathbf{W}_j^{\triangleright}\) of the SLFN. When the number of neurons in the hidden layer \(h_{j}\) is sufficiently large, there exist weight matrices \(\mathbf{W}_j\) and \(\mathbf{W}_j^{\triangleright}\) such that the target provided in \eqref{eq:1} can be approximated by the network whose output is given in \eqref{eq:rbf_output}
\begin{eqnarray}
\boldsymbol{y}(k)&=& \boldsymbol{y}_{\mathcal{NN}}(\mathbf{W}_j,\mathbf{W}_j^{\vartriangleright},\boldsymbol{z}_j(k))\nonumber\\
&=&\boldsymbol{\sigma}\big(\mathbf{W}_j^{\triangleright} \boldsymbol{\phi}_{j}(\mathbf{W}_j \boldsymbol{z}_j(k))\big)  \label{eq:yk_2}
\end{eqnarray}
As \(\mathbf{W}_j\) and \(\mathbf{W}_j^{\triangleright}\) are unknown, they are updated incrementally by two learning laws. Let \(\hat{\mathbf{W}}_j(k)\) and \(\hat{\mathbf{W}}_j^{\triangleright}(k)\) denote the estimated weight matrices \(\mathbf{W}_j\) and \(\mathbf{W}_j^{\triangleright}\) at the \(k^{\text{th}}\) step of learning, the estimated output vector \(\hat{\boldsymbol{y}}(k)\) at the \(k^{\text{th}}\) step is constructed as the output of the SLFN when its weights are set at \(\hat{\mathbf{W}}_j(k)\) and \(\hat{\mathbf{W}}_j^{\triangleright}(k)\)
\begin{eqnarray}
\hat{\boldsymbol{y}}(k)&=&\boldsymbol{y}_{\mathcal{NN}}(\hat{\mathbf{W}}_j(k),\hat{\mathbf{W}}_j^{\vartriangleright}(k),\boldsymbol{z}_j(k))\nonumber\\
&=&\boldsymbol{\sigma}\big(\hat{\mathbf{W}}_j^{\triangleright}(k) \boldsymbol{\phi}_{j}(\hat{\mathbf{W}}_j(k) \boldsymbol{z}_j(k))\big) \label{eq:est_yk_2}
\end{eqnarray}
The output estimation error at the \(k^{\text{th}}\) step is defined as\\ \(\boldsymbol{e}(k)=\boldsymbol{y}(k)-\hat{\boldsymbol{y}}(k)\). Hence,
\begin{eqnarray}
\boldsymbol{e}(k)
&=&\boldsymbol{y}_{\mathcal{NN}}(\mathbf{W}_j,\mathbf{W}_j^{\vartriangleright},\boldsymbol{z}_j(k))\nonumber\\
&&{-\;}\boldsymbol{y}_{\mathcal{NN}}(\hat{\mathbf{W}}_j(k),\hat{\mathbf{W}}_j^{\vartriangleright}(k),\boldsymbol{z}_j(k)) \label{eq:mlfn_general_er_2}\\
&=&\boldsymbol{\sigma}\big(\mathbf{W}_j^{\triangleright} \boldsymbol{\phi}_{j}(\mathbf{W}_j \boldsymbol{z}_j(k))\big) \nonumber\\
&&{-\;}  \boldsymbol{\sigma}\big(\hat{\mathbf{W}}_j^{\triangleright}(k) \boldsymbol{\phi}_{j}(\hat{\mathbf{W}}_j(k) \boldsymbol{z}_j(k))\big)  \label{eq:err_2}
\end{eqnarray}
Let 
\begin{equation}
\boldsymbol{\delta}(k) \triangleq \mathbf{W}_j^{\triangleright} \boldsymbol{\phi}_{j}(\mathbf{W}_j \boldsymbol{z}_j(k)) - \hat{\mathbf{W}}_j^{\triangleright}(k) \boldsymbol{\phi}_{j}(\hat{\mathbf{W}}_j(k) \boldsymbol{z}_j(k)) \label{eq:delta_2}
\end{equation}  
which can be expressed as
\begin{eqnarray} 
\boldsymbol{\delta}(k) 
&=&\hat{\mathbf{W}}_j^{\triangleright}(k)\Delta \boldsymbol{\phi}_{j}(k) + \Delta\mathbf{W}_j^{\triangleright}(k)\hat{\boldsymbol{\phi}}_{j}(k)\nonumber\\
&&{+\;} \Delta\mathbf{W}_j^{\triangleright}(k) \Delta \boldsymbol{\phi}_{j}(k) \label{eq:deltak_2}
\end{eqnarray}
where \(\Delta \boldsymbol{\phi}_{j}(k) \triangleq  \boldsymbol{\phi}_{j}(\mathbf{W}_j \boldsymbol{z}_j(k)) - \boldsymbol{\phi}_{j}(\hat{\mathbf{W}}_j(k) \boldsymbol{z}_j(k))\), \(\hat{\boldsymbol{\phi}}_{j}(k) \triangleq \boldsymbol{\phi}_{j}(\hat{\mathbf{W}}_j(k) \boldsymbol{z}_j(k))\) and \(\Delta\mathbf{W}_j^{\triangleright}(k) \triangleq {\mathbf{W}}_j^{\triangleright} -  \hat{\mathbf{W}}_j^{\triangleright}(k) \). \vspace{-2mm}

Consequently, the incremental learning laws (two-layer update) to update the estimated weight based on the output estimation error \(\boldsymbol{e}(k)\) are proposed as:
\begin{eqnarray}
\hat{\mathbf{W}}_j^{\triangleright}(k+1)=\hat{\mathbf{W}}_j^{\triangleright}(k)+ \alpha_1 \mathbf{L}(k) \boldsymbol{e}(k)  \hat{\boldsymbol{\phi}}_{j}^T(k) \label{eq:updL_2layer_a} \\
\hat{\mathbf{W}}_j(k+1)=\hat{\mathbf{W}}_j(k)+ \alpha_0 \mathbf{P}(k) \boldsymbol{e}(k)  \boldsymbol{z}_j^T(k) \label{eq:updL_2layer_b} 
\end{eqnarray}
where \(\alpha_1\) and \(\alpha_0\) are positive scalars, \(\mathbf{L}(k) \in \mathbb{R}^{p \times p}\) is a positive diagonal matrix, \({\mathbf{P}}(k) \in \mathbb{R}^{h_{j} \times p}\) is a matrix depending on the learning step \(k\). \vspace*{-2mm}

\noindent In \eqref{eq:updL_2layer_a}, let \(\boldsymbol{w}_{j,i}^{\triangleright}\) denote the \(i^{\text{th}}\) column vector of matrix \(\mathbf{W}_j^{\triangleright}\),  \(\hat{\boldsymbol{w}}_{j,i}^{\triangleright}(k)\) the  \(i^{\text{th}}\) column vector of \(\hat{\mathbf{W}}_j^{\triangleright}(k)\) and \(\hat{\phi}_{j,i}(k)\) the \(i^{\text{th}}\) element of vector \(\boldsymbol{\phi}_{j}(k)\).
In \eqref{eq:updL_2layer_b}, let \(\boldsymbol{w}_{j,\imath}\) denote the \(\imath^{\text{th}}\) column vector of matrix \(\mathbf{W}_j\),  \(\hat{\boldsymbol{w}}_{j,\imath}(k)\) the  \(\imath^{\text{th}}\) column vector of \(\hat{\mathbf{W}}_j(k)\) and \(z_{j,\imath}(k)\) the \(\imath^{\text{th}}\) element of the vector \(\boldsymbol{z}_j(k)\).
The update laws \eqref{eq:updL_2layer_a} and \eqref{eq:updL_2layer_b} can be rewritten in vector form as:
\begin{eqnarray}
\hat{\boldsymbol{w}}_{j,i}^{\triangleright}(k+1)=\hat{\boldsymbol{w}}_{j,i}^{\triangleright}(k)+  \alpha_1 \hat{\phi}_{j,i}(k) \mathbf{L}(k)\boldsymbol{e}(k)  \label{eq:updL_2layer_av}\\
\hat{\boldsymbol{w}}_{j,\imath}(k+1)=\hat{\boldsymbol{w}}_{j,\imath}(k)+ \alpha_0 z_{j,\imath}(k) {\mathbf{P}}(k) \boldsymbol{e}(k)  \label{eq:updL_2layer_bv}  
\end{eqnarray}
To show the convergence, we define an objective function given by \vspace*{-2mm}
\begin{eqnarray}
V(k)&=& \frac{1}{\alpha_1}\sum_{i=1}^{h_{j}}{\Delta \boldsymbol{w}_{j,i}^{\triangleright T}(k) \Delta \boldsymbol{w}_{j,i}^{\triangleright}(k)} \nonumber\\
&&{+\;} \frac{1}{\alpha_0} \sum_{\imath=1}^{h_{j-1}}{\Delta \boldsymbol{w}_{j,\imath}^T(k) \Delta \boldsymbol{w}_{j,\imath}(k)}  \label{eq:Vk_2}
\end{eqnarray}
where \(\Delta \boldsymbol{w}_{j,i}^{\triangleright}(k) = \boldsymbol{w}_{j,i}^{\triangleright} -\hat{\boldsymbol{w}}_{j,i}^{\triangleright}(k)\) and \(\Delta \boldsymbol{w}_{j,\imath}(k) = \boldsymbol{w}_{j,\imath} -\hat{\boldsymbol{w}}_{j,\imath}(k)\).
From \eqref{eq:updL_2layer_av} and \eqref{eq:updL_2layer_bv}, the objective function at the \((k+1)^\text{th}\) step can be written as
\begin{align}
V(k+1)  = & \frac{1}{\alpha_1}\sum_{i=1}^{h_{j}}{\Delta \boldsymbol{w}_{j,i}^{\triangleright T}(k+1) \Delta \boldsymbol{w}_{j,i}^{\triangleright}(k+1)} \nonumber \\
&{+\;} \frac{1}{\alpha_0} \sum_{\imath=1}^{h_{j-1}}{\Delta \boldsymbol{w}_{j,\imath}^T(k+1) \Delta \boldsymbol{w}_{j,\imath}(k+1)} 
\end{align}
Using \eqref{eq:updL_2layer_av} and \eqref{eq:updL_2layer_bv}, a change of the objective function value when the training step goes from \(k^\text{th}\) to \((k+1)^\text{th}\) can therefore be derived as
\begin{eqnarray}
\Delta V(k) &=&V(k+1)-V(k) \nonumber\\
&=&-\; \hat{\boldsymbol{\phi}}_{j}^T(k)\Delta\mathbf{W}_j^{\triangleright T}(k) \mathbf{L}(k)\boldsymbol{e}(k) \nonumber\\
&&{-\;} \boldsymbol{z}_j^T(k) \Delta\mathbf{W}_j^T(k) \mathbf{P}(k)\boldsymbol{e}(k)\nonumber
\end{eqnarray}
\begin{eqnarray}
&&{-\;} \boldsymbol{e}^T(k)\mathbf{L}^T(k)\Delta\mathbf{W}_j^{\triangleright}(k)\hat{\boldsymbol{\phi}}_{j}(k) \nonumber\\
&&{-\;} \boldsymbol{e}^T(k)\mathbf{P}^T(k) \Delta\mathbf{W}_j(k)\boldsymbol{z}_j(k)  \nonumber\\
&&{+\;}\boldsymbol{e}^T(k) \Big(\alpha_1\hat{\mu}_{j}(k)\mathbf{L}^T(k)\mathbf{L}(k) \nonumber\\
&&{+\;}  \alpha_0\rho_{j}(k) \mathbf{P}^T(k) \mathbf{P}(k)\Big) \boldsymbol{e}(k)
\label{eq:deltaV_2}
\end{eqnarray}
{with \(\hat{\mu}_{j}(k)\triangleq\sum_{i=1}^{h_{j}}\hat{\phi}_{j,i}^2(k)\) and \(\rho_{j}(k)\triangleq\sum_{\imath=1}^{h_{j-1}} z_{j,\imath}^2(k)\).}
From \eqref{eq:deltak_2} , we have
\begin{eqnarray}
&\Delta\mathbf{W}_j^{\triangleright}(k)\hat{\boldsymbol{\phi}}_{j}(k) \nonumber\\
&=
\boldsymbol{\delta}(k) - \hat{\mathbf{W}}_j^{\triangleright}(k)\Delta \boldsymbol{\phi}_{j}(k) -  \Delta\mathbf{W}_j^{\triangleright}(k) \Delta \boldsymbol{\phi}_{j}(k)
\end{eqnarray}
Next, substituting into \eqref{eq:deltaV_2} gives
\begin{eqnarray}
\Delta V(k) 
&=&{-\;} \boldsymbol{\delta}^T(k)\mathbf{L}(k)\boldsymbol{e}(k)- \boldsymbol{e}^T(k)\mathbf{L}^T(k)\boldsymbol{\delta}(k) \nonumber \\ 
&&{+\;}\Big(\Delta{\boldsymbol{\phi}_{j}}^T(k)\hat{\mathbf{W}}_j^{\triangleright T}(k)  \nonumber\\
&&{-\;} \boldsymbol{z}_j^T(k) \Delta\mathbf{W}_j^T(k) \mathbf{P}(k) \mathbf{L}^{-1}(k)\Big)\mathbf{L}(k)\boldsymbol{e}(k) \nonumber\\
&&{+\;}\boldsymbol{e}^T(k)\mathbf{L}^T(k) \Big( \hat{\mathbf{W}}_j^{\triangleright}(k)\Delta{\boldsymbol{\phi}_{j}}(k) \nonumber\\
&&{-\;} \mathbf{L}^{-T}(k)\mathbf{P}^T(k) \Delta\mathbf{W}_j(k)\boldsymbol{z}_j(k)   \Big) \nonumber\\
&&{+\;}\boldsymbol{e}^T(k) \Big(\alpha_1 \hat{\mu}_{j}(k)\mathbf{L}^T(k)\mathbf{L}(k)  \nonumber\\
&&{+\;} \alpha_0 \rho_{j}(k) \mathbf{P}^T(k)\mathbf{P}(k) \Big) \boldsymbol{e}(k) \nonumber \\
&&{+\;}\Delta \boldsymbol{\phi}_{j}^T(k) \Delta\mathbf{W}_j^{\triangleright T}(k)\mathbf{L}(k)\boldsymbol{e}(k) \nonumber\\
&&{+\;} \boldsymbol{e}^T(k)\mathbf{L}^T(k)\Delta\mathbf{W}_j^{\triangleright}(k) \Delta \boldsymbol{\phi}_{j}(k)
\label{eq:deltaV_3}
\vspace*{-2mm}
\end{eqnarray}
\noindent After the pre-training phase, the errors are sufficiently small and hence the last two terms which are of \(O^3 \) are negligible as compared to the other terms which are of \(O^2\).
Also, let the matrix \(\mathbf{P}(k)\) be chosen so that \(\boldsymbol{\xi}(k) \triangleq  \hat{\mathbf{W}}_j^{\triangleright}(k)\Delta{\boldsymbol{\phi}_{j}}(k) - \mathbf{L}^{-T}(k)\mathbf{P}^T(k) \Delta\mathbf{W}_j(k)\boldsymbol{z}_j(k)  \) is zero or sufficiently small, then the equation \eqref{eq:deltaV_3} becomes
\begin{eqnarray}
\Delta V(k) &=&-{\;}\boldsymbol{\delta}^T(k)\mathbf{L}(k)\boldsymbol{e}(k)- \boldsymbol{e}^T(k)\mathbf{L}^T(k)\boldsymbol{\delta}(k) \nonumber \\
&&{+\;}\boldsymbol{e}^T(k) \Big(\alpha_1 \hat{\mu}_{j}(k)\mathbf{L}^T(k)\mathbf{L}(k)  \nonumber \\
&&{+\;} \alpha_0 \rho_{j}(k) \mathbf{P}^T(k)\mathbf{P}(k)\Big) \boldsymbol{e}(k) 
\label{eq:deltaV_5}
\vspace*{-2mm}
\end{eqnarray}
Similarly to the one-layer update, if the activation functions in \(\boldsymbol{\sigma}\) are monotonically increasing and their derivatives are bounded by \(f_{\sigma}\), comparing between \(\boldsymbol{e}(k)\) in \eqref{eq:err_2} and \(\boldsymbol{\delta}(k)\) in \eqref{eq:delta_2}, { then:} \vspace*{-3mm}
\begin{itemize}
	\item[i,]the corresponding elements of \( \boldsymbol{e}(k)\)  and \(\boldsymbol{\delta}(k)\)  have the same sign, i.e.
	\begin{equation}
	e_{i}(k)\delta_{i}(k) \geq 0,{\;} \forall i = 1..p\label{con_1_2}
	\end{equation}
	\item[ii,]the absolute values of the elements of \(\boldsymbol{e}(k)\)  are less than or equal to \( f_{\sigma}\) times the corresponding elements of
	\(\boldsymbol{\delta}(k)\), i.e.
	\begin{equation}
	|{e}_{i}(k)| \leq  f_{\sigma}|{\delta}_{i}(k)|,{\;} \forall i = 1..p \label{con_2_2}
	\end{equation} 
\end{itemize}
From the properties stated in \eqref{con_1_2}, \eqref{con_2_2}, the following inequality can be assured
\begin{eqnarray}
\Delta V(k) &\leq&-\frac{2}{f_{\sigma}}\boldsymbol{e}^T(k)\mathbf{L}(k)\boldsymbol{e}(k)   \nonumber 
\end{eqnarray}
\begin{eqnarray}
&&{+\;}\boldsymbol{e}^T(k) \Big(\alpha_1 \hat{\mu}_{j}(k)\mathbf{L}^T(k)\mathbf{L}(k) \nonumber \\
&&{+\;} \alpha_0 \rho_{j}(k) \mathbf{P}^T(k)\mathbf{P}(k) \Big) \boldsymbol{e}(k) \label{eq:deltaV_final_2} 
\end{eqnarray}
When \(\mathbf{L}(k)\) is chosen such that
\begin{eqnarray}
&&\frac{2}{f_{\sigma}}\mathbf{L}(k)-\Big(\alpha_1 \hat{\mu}_{j}(k)\mathbf{L}^T(k)\mathbf{L}(k) \nonumber \\
&&{+\;} \alpha_0 \rho_{j}(k) \mathbf{P}^T(k)\mathbf{P}(k) \Big) >0 \label{eq:con_L_2} 
\end{eqnarray}
then \( \Delta V(k) \) is non-positive for any \(\boldsymbol{e}(k)\). That means, the value of the objective function is a non-increasing
sequence \(V(k+1)<V(k)\). Moreover, since the function \(V(k)\) is non-negative, which means that it is bounded from below, we have \( \Delta V(k)\) converges as \(k\) increases. Thus, from \eqref{eq:deltaV_final_2}, \( \boldsymbol{e}(k)\) converges as \(k\) increases. \vspace*{-2mm}

However, to achieve
\begin{flalign}
\boldsymbol{\xi}(k) &=	\hat{\mathbf{W}}_j^{\triangleright}(k)\Delta{\boldsymbol{\phi}_{j}}(k) - \mathbf{L}^{-T}(k)\mathbf{P}^T(k) \Delta\mathbf{W}_j(k)\boldsymbol{z}_j(k)\nonumber\\ &\approx 0
\end{flalign}
 is not straightforward as it depends on the choices of \(\boldsymbol{\phi}_{j}\). Let us now analyze the choice of \(\mathbf{P}(k)\).
\\
\indent \textbf{Discussions:} Setting \(\mathbf{P}(k) = \mathbf{\Theta}^T(k)\hat{\mathbf{W}}_j^{\triangleright T}(k)\mathbf{L}(k)\), where \(\mathbf{\Theta}(k) \in \mathbb{R}^{h_{j} \times h_{j}}\), we have
\begin{flalign}
\boldsymbol{\xi}(k)  = \hat{\mathbf{W}}_j^{\triangleright}(k)\left(\Delta{\boldsymbol{\phi}_{j}}(k)-\mathbf{\Theta}(k)\Delta\mathbf{W}_j(k)\boldsymbol{z}_j(k)\right)
\end{flalign} 
To achieve \(\boldsymbol{\xi}(k) \approx 0\), the matrix \(\mathbf{\Theta}(k)\) should be chosen such that \(\Delta{\boldsymbol{\phi}_{j}}(k)-\mathbf{\Theta}(k)\Delta\mathbf{W}_j(k)\boldsymbol{z}_j(k) \approx 0 \), or
\begin{eqnarray}
\Delta{\boldsymbol{\phi}_{j}}(k) &=& \boldsymbol{\phi}_{j}(\mathbf{W}_j \boldsymbol{z}_j(k)) - \boldsymbol{\phi}_{j}(\hat{\mathbf{W}}_j(k) \boldsymbol{z}_j(k)) \nonumber\\
&\approx& \mathbf{\Theta}(k)\Delta\mathbf{W}_j(k)\boldsymbol{z}_j(k) \label{eq:term_condition} 
\end{eqnarray} 
Noting that the activation functions in the vector \(\boldsymbol{\phi}_{j}\) act element-wisely, therefore it is clearer to look at each element of the vector \(\Delta{\boldsymbol{\phi}_{j}}(k)\). Its \(i^\text{th}\) element is denoted as \(\Delta{\phi}_{j,i}(k) = {\phi}_{j,i}({\boldsymbol{w}}_{j,ri} \boldsymbol{z}_j(k)) - {\phi}_{j,i}( \hat{\boldsymbol{w}}_{j,ri}(k) \boldsymbol{z}_j(k))\) where  \({\boldsymbol{w}}_{j,ri},\hat{\boldsymbol{w}}_{j,ri}(k)\) are the \(i^\text{th}\) rows of matrices \({\mathbf{W}}_j,\hat{\mathbf{W}}_j(k)\), respectively.
Now let us look at the choice of \(\mathbf{\Theta}(k)\) with different activation functions \(\boldsymbol{\phi}_{j}\).

It is interesting to note that if the activation functions in the vector \(\boldsymbol{\phi}_{j}\) are chosen as the ReLUs, which are defined as \(\phi(x) = x \text{ if } x > 0, \text{ and } \phi(x) = 0 \text{ if otherwise}\). Hence, we have \(\Delta{\phi}_{j,i}(k) =\) \vspace*{-2mm}
\begin{itemize}
\item \(\Delta{\boldsymbol{w}}_{j,ri}(k) \boldsymbol{z}_j(k)\) if \({\boldsymbol{w}}_{j,ri} \boldsymbol{z}_j(k) \geq 0\) and \(\hat{\boldsymbol{w}}_{j,ri}(k) \boldsymbol{z}_j(k) \geq 0\);
\item  \(0\) if \({\boldsymbol{w}}_{j,ri} \boldsymbol{z}_j(k) \leq 0\) and \(\hat{\boldsymbol{w}}_{j,ri}(k) \boldsymbol{z}_j(k) \leq 0\);
\item \({\boldsymbol{w}}_{j,ri} \boldsymbol{z}_j(k)\) if \({\boldsymbol{w}}_{j,ri} \boldsymbol{z}_j(k) \geq 0\) and \(\hat{\boldsymbol{w}}_{j,ri}(k) \boldsymbol{z}_j(k) \leq 0\);
\item \(-\hat{\boldsymbol{w}}_{j,ri}(k) \boldsymbol{z}_j(k)\) if \({\boldsymbol{w}}_{j,ri} \boldsymbol{z}_j(k) \leq 0\) and \(\hat{\boldsymbol{w}}_{j,ri}(k) \boldsymbol{z}_j(k) \geq 0\).
\vspace{-2mm}
\end{itemize}
We consider \({\boldsymbol{w}}_{j,ri}\boldsymbol{z}_j(k)\) and  \(\hat{\boldsymbol{w}}_{j,ri}(k)\boldsymbol{z}_j(k)\) which are close to each other after the pre-training phase. Hence, in the last two cases above where they are of opposite signs, they should be sufficiently small. Therefore, it can be considered that for these last two cases \(\Delta{\phi}_{j,i}(k) \approx 0\). 
Hence, the matrix \(\mathbf{\Theta}(k)\) in \eqref{eq:term_condition} can be chosen as a diagonal matrix \(\text{diag}\{\theta_1(k), \theta_2(k), \ldots, \theta_{h_j}(k)\}\) where the diagonal elements are
\begin{eqnarray}
\theta_{i}(k)=\left\{
\begin{array}{ll}
1 \text{ if } \hat{\boldsymbol{w}}_{j,ri}(k) \boldsymbol{z}_j(k) \geq 0\\
0 \text{ otherwise } 
\end{array}
\right. \label{eq:relu_theta}
\end{eqnarray}
Looking back at \eqref{eq:relu_theta}, the value of \(\theta_i(k)\) is actually the derivative of the ReLU function
\begin{eqnarray}
\theta_i(k) = \hat{{\phi}}'_{j,i}(k) = \frac{d\phi_{j,i}(x(k))}{dx(k)}\Big|_{x(k)=\hat{\boldsymbol{w}}_{j,ri}(k) \boldsymbol{z}_j(k)} \label{eq:theta_general} 
\end{eqnarray}
Generalizing to any differentiable function, we set \(\mathbf{\Theta}(k)= \mathbf{\Phi}'_j(k)\)  with \(\mathbf{\Phi}'_j(k)=\text{diag}\{\theta_1(k)\), \( \theta_2(k)\), \(\ldots \), \(\theta_{h_j}(k)\}\) where \(\theta_i(k)\) is defined in \eqref{eq:theta_general}. 
With this first order approximation, we have
\begin{eqnarray}
\boldsymbol{\phi}_{j}(\mathbf{W}_j \boldsymbol{z}_j(k)) - \boldsymbol{\phi}_{j}(\hat{\mathbf{W}}_j(k) \boldsymbol{z}_j(k)) 
\approx \mathbf{\Phi}'_j(k)\Delta\mathbf{W}_j(k)\boldsymbol{z}_j(k) \nonumber
\end{eqnarray}
Hence, equation \eqref{eq:term_condition} is satisfied. Now, replace the matrix \(\mathbf{P}(k)\) into \eqref{eq:updL_2layer_b} and rewrite the full update law in \eqref{eq:updL_2layer_a} and \eqref{eq:updL_2layer_b} as follows
\begin{flalign}
&\hat{\mathbf{W}}_j^{\triangleright}(k+1)=\hat{\mathbf{W}}_j^{\triangleright}(k)+ \alpha_1 \mathbf{L}(k) \boldsymbol{e}(k)  \hat{\boldsymbol{\phi}}_{j}^T(k) \label{eq:updL_2layer_a_final} \\
&\hat{\mathbf{W}}_j(k+1)=\hat{\mathbf{W}}_j(k)+ \alpha_0 \mathbf{\Phi}_j'^T(k)\hat{\mathbf{W}}_j^{\triangleright T}(k)\mathbf{L}(k) \boldsymbol{e}(k)  \boldsymbol{z}_j^T(k) \label{eq:updL_2layer_b_final} 
\end{flalign}
It can be seen that these update laws are similar to the first order gradient descent when the activation functions at the output layer are linear. \vspace*{-2.5mm}

The complete condition \eqref{eq:con_L_2} can now be written as
\begin{flalign}
&\frac{2}{f_{\sigma}}\mathbf{L}(k)-\Big(\alpha_1 \hat{\mu}_{j}(k)\mathbf{L}^T(k)\mathbf{L}(k) \nonumber \\
&{+\;} \alpha_0 \rho_{j}(k) \mathbf{L}^T(k) \hat{\mathbf{W}}_j^{\triangleright}(k) \mathbf{\Phi}_j'(k) \nonumber\\
&{\;\;\;\;\;\;\;\;\;\;\;\;\;\;\; \times \;}\mathbf{\Phi}_j'^T(k)\hat{\mathbf{W}}_j^{\triangleright T}(k)\mathbf{L}(k) \Big) >0 \label{eq:con_L_2_final} \vspace{-2mm}
\end{flalign} 
\noindent{The above condition suggests that the training at the fine-tuning stage should be done with small gains by setting small values of \(\alpha_0\) and \(\alpha_1\) so as to ensure convergence. Similarly, the gain matrix \(\mathbf{L}(k)\) can be initialized to any arbitrary value, and then adjusted automatically by monitoring the condition in each update and reducing it if necessary. }\vspace{-2.5mm}

\noindent{\textbf{Remark 5}: In the analysis of the two-layer update algorithm {in the FPL method}, condition \eqref{eq:term_condition} has been used for proving the convergence. This condition suggests that the two-layer update algorithm should be used for fine tuning so as to ensure convergence.  That is, a pre-training phase should first be conducted by using the inverse layer-wise learning so that the error is {relatively} small after pre-training. Therefore, based on the convergence analysis, we propose in this paper that the training should be done in two phases: pre-training and fine-tuning, and the condition for the gain matrix in \eqref{eq:con_L_2_final} can be used in the latter phase of training of the network. It is also worth noting that such condition is not required in the analysis of the inverse layer-wise learning in section \ref{sec: main_1}.}
\vspace{-2.5mm}

\noindent{\textbf{Remark 6}: In case of having an insufficient number of hidden neurons and/or in presence of measurement noise, {similar to Remark 4}, it can be shown that if
	\begin{align} 
	&\medmath{\Vert {\boldsymbol{e}}(k) \Vert  \geq  \frac{b}{2\left( \frac{2L_{m}}{f_{\sigma M}}-d_M L_{M}^2\right)}} \Bigg[\medmath{\frac{4L_{m}}{f_{\sigma M}}+\frac{2L_{M}}{f_{\sigma m}}}\nonumber\\
	 &\medmath{+\sqrt{\frac{8L_{m}}{f_{\sigma M}}d_M L_{M}^2+ \frac{8L_{M}}{f_{\sigma m}}\left( \frac{4L_{m}}{f_{\sigma M}}-d_M L_{M}^2\right)+\left(\frac{2L_{M}}{f_{\sigma m}}\right)^2}}\Bigg] \label{eq:bound_2lay} 
	\end{align}
	\vspace{-3mm}
	\begin{flalign}
	&\text{and\;\;\;\;\;\;}\frac{2L_{m}}{f_{\sigma M}}-d_M L_{M}^2 > 0&   
	\end{flalign}
	then \(\Delta V(k) \leq 0\). 
	Here \(b\) is the upper bound of the NN approximation error and measurement noise, and \(d_M\) is a positive constant such that
	\begin{flalign}
	&\boldsymbol{e}^T(k) \Big(\alpha_1 \hat{\mu}_{j}(k)\mathbf{L}^T(k)\mathbf{L}(k)  \nonumber\\
	&+ \alpha_0 \rho_{j}(k) \mathbf{P}^T(k)\mathbf{P}(k)\Big) \boldsymbol{e}(k) \leq d_M L_{M}^2   \Vert {\boldsymbol{e}}(k) \Vert^2 \vspace{-3mm}
	\end{flalign} 
	Therefore, there exists an ultimate bound such that the error always stay within the bound after reaching it. Noting {from \eqref{eq:bound_2lay}} that the bound would tend to zero when the NN approximation error and noise tend to zero.}\vspace{-2.5mm}

\subsection{Summary of Forward Progressive Learning Algorithm} \label{sec:slfn_sum} \vspace*{-2.5mm}
The overall algorithm for learning of each SLFN in the FPL is described as follows: \vspace*{-2.5mm}
\begin{enumerate}
	\item [(i)] The first phase: Pre-train the SLFN following the steps in subsection \ref{sec:pre-train}. 
	\item [(ii)] The second phase: Fine-tuning of the SLFN:
	\begin{itemize}
	\item [(a)] For each data sample \((\boldsymbol{z}_j(k),\boldsymbol{y}(k))\) in the training set:
	\begin{itemize}
	\item [(1)] Calculate \(\hat{\boldsymbol{y}}(k)\) using \eqref{eq:est_yk_2}.
	\item [(2)] Calculate \(\boldsymbol{e}(k)=\boldsymbol{y}(k)-\hat{\boldsymbol{y}}(k)\).
	\item [(3)] Update the weight matrices using laws \eqref{eq:updL_2layer_a_final} and \eqref{eq:updL_2layer_b_final} where \(\mathbf{L}(k)\) should satisfy \eqref{eq:con_L_2_final}. 
	\end{itemize}
	\item [(b)] Move to the next training sample and repeat (a) (steps (1) to (3)) until all the samples in the training set have been used.
	\item [(c)] When all the data samples in the training set have been used \(\to\)  finish 1 loop. The training can take more than 1 loop if necessary.
	\end{itemize}
\vspace*{-2.5mm}
\end{enumerate}

\section{Online Kinematic Control of Robot Manipulators} \label{sec: main_3}\vspace*{-2.5mm}
In this section, we show how the results can be adapted for online learning of robot kinematics without any modeling. \vspace*{-2.5mm}

{At each sampling time instance \(k\), the rate of change of joint variables \(\dot{\boldsymbol{q}}(k)\) is related to the rate of change of position and orientation of the end effector in sensory space \(\dot{\boldsymbol{x}}(k)\) as
\begin{flalign}
&\dot{\boldsymbol{x}}(k)= \mathbf{J}(\boldsymbol{q}(k))\dot{\boldsymbol{q}}(k) &\label{eq:jacob_all}
\end{flalign}
where \(\mathbf{J}(\boldsymbol{q}(k))\) is the overall Jacobian matrix from joint space to sensory task space.} \vspace{-2mm}

\noindent The relationship in equation \eqref{eq:jacob_all} can be approximated by a multilayer network whose output is given in equation \eqref{eq:mlfn_output_short}
\begin{flalign}
&\dot{{\boldsymbol{x}}}(k) = {\mathbf{J}}(\boldsymbol{q}(k))\dot{\boldsymbol{q}}(k) =  \boldsymbol{y}_{\mathcal{NN}}(\mathbf{W}_j|_{j=1}^{n},\boldsymbol{q}(k),\dot{\boldsymbol{q}}(k)) \label{eq:xk_dot}
\end{flalign}
It can be seen that the learning algorithms in section \ref{sec: main_1} and section \ref{sec: main_2} can be directly applied for offline learning by setting \(\boldsymbol{y} = \dot{\boldsymbol{x}}\) and inputting to the network \(\boldsymbol{q},\dot{\boldsymbol{q}}\). However, for online robot control, a desired trajectory is specified and hence the learning algorithms need to be adapted for online learning purpose. \vspace*{-2mm}

\noindent The estimated output of the network at the \(k^\text{th}\) step of online learning is given as
\begin{eqnarray}
\dot{\hat{\boldsymbol{x}}}(k) &=&  \hat{\mathbf{J}}(\boldsymbol{q}(k),\hat{\mathbf{W}}_{\Sigma}(k))\dot{\boldsymbol{q}}(k)  \nonumber\\
&=&  \boldsymbol{y}_{\mathcal{NN}}(\hat{\mathbf{W}}_j(k)|_{j=1}^{n},\boldsymbol{q}(k),\dot{\boldsymbol{q}}(k)) \label{eq:xk_dot_hat}
\end{eqnarray}
{where \(\hat{\mathbf{W}}_{\Sigma}(k)\) stands for all the estimated weight matrices \(\hat{\mathbf{W}}_j(k)\) for \(j=1,2,\ldots,n\).} \vspace*{-2mm}

Let the reference joint velocity \(\dot{\boldsymbol{q}}\) based on the sensory task space feedback be proposed as follows
\begin{eqnarray}
\dot{\boldsymbol{q}}(k) &=& \hat{\mathbf{J}}^\dagger(\boldsymbol{q}(k),\hat{\mathbf{W}}_{\Sigma}(k))(\dot{\boldsymbol{x}}_d(k) -\alpha\Delta \boldsymbol{x}(k)) \label{eq:control_qre}
\end{eqnarray}
where \(\hat{\mathbf{J}}^\dagger(\boldsymbol{q}(k),\hat{\mathbf{W}}_{\Sigma}(k))\) is the pseudoinverse matrix of the estimated Jacobian \(\hat{\mathbf{J}}(\boldsymbol{q}(k),\hat{\mathbf{W}}_{\Sigma}(k))\); \(\alpha\) is a positive scalar; \(\Delta \boldsymbol{x}(k) = \boldsymbol{x}(k) - \boldsymbol{x}_d(k)\); \(\boldsymbol{x}_d(k)\) and  \(\dot{\boldsymbol{x}}_d(k)\), respectively, are the desired position and velocity of the end effector in the sensory task space.
Premultiplying \eqref{eq:control_qre} by \(\hat{\mathbf{J}}(\boldsymbol{q}(k),\hat{\mathbf{W}}_{\Sigma}(k))\) gives
\begin{eqnarray}
\hat{\mathbf{J}}(\boldsymbol{q}(k),\hat{\mathbf{W}}_{\Sigma}(k))\dot{\boldsymbol{q}}(k) = \dot{\boldsymbol{x}}_d(k) -\alpha\Delta \boldsymbol{x}(k) \label{eq:premul_qre}
\end{eqnarray}
Subtracting \eqref{eq:xk_dot} and \eqref{eq:premul_qre} gives
\begin{eqnarray}
&&{\mathbf{J}}(\boldsymbol{q}(k))\dot{\boldsymbol{q}}(k)-\hat{\mathbf{J}}(\boldsymbol{q}(k),\hat{\mathbf{W}}_{\Sigma}(k))\dot{\boldsymbol{q}}(k) \nonumber\\
&=& \dot{{\boldsymbol{x}}}(k)-\dot{\boldsymbol{x}}_d(k) +\alpha\Delta \boldsymbol{x}(k) =\Delta \dot{\boldsymbol{x}}(k) +\alpha\Delta \boldsymbol{x}(k) \label{eq:dyna_erro}
\end{eqnarray}
Let \(\boldsymbol{\varepsilon}(k) \triangleq \Delta \dot{\boldsymbol{x}}(k) +\alpha\Delta \boldsymbol{x}(k)\) be the online feedback error in online learning, from \eqref{eq:xk_dot} and \eqref{eq:xk_dot_hat} we have
\begin{eqnarray}
\boldsymbol{\varepsilon}(k) &=& \boldsymbol{y}_{\mathcal{NN}}(\mathbf{W}_j|_{j=1}^{n},\boldsymbol{q}(k),\dot{\boldsymbol{q}}(k)) \nonumber \\
&&{-\;}\boldsymbol{y}_{\mathcal{NN}}(\hat{\mathbf{W}}_j(k)|_{j=1}^{n},\boldsymbol{q}(k),\dot{\boldsymbol{q}}(k)) \label{eq:dyna_err}
\end{eqnarray}
{Hence, the online feedback error \(\boldsymbol{\varepsilon}(k)\) here is equal to the output estimation error as seen in \eqref{eq:mlfn_general_er_2}}.  By using \(\boldsymbol{\varepsilon}(k)\) for the update laws in section \ref{sec: main_2} to train the network, {the convergence of this error can be guaranteed.} \vspace*{-2mm}

{In most robot systems, the joint positions are measured by encoders and the joint velocities are obtained from differentiation of the positions.  Similarly, for task space sensory feedback control, the positions of the end-effector are measured by a sensor such as a camera and the velocities are obtained from differentiation.} \vspace*{-2mm}

\section{Case Studies} \label{sec: case_study} \vspace{-2mm}
In this section, we present {four} case studies to illustrate the performance of the proposed learning algorithms. { The first experiment is on a {multivariable} function approximation problem. Next, two classification problems with the classical MNIST and CIFAR-10 databases are presented. {Finally,} a regression problem with an online tracking control task for a UR5e manipulator {is also presented}. \vspace*{-2mm}

\subsection{Nonlinear function approximation} \label{sec: sinexp} \vspace{-2mm}
We considered the nonlinear function:
\begin{equation}
y(\boldsymbol{z}) = e^{0.5z_1} sin(\frac{\pi}{4}z_2)+0.2(z_1+z_2)^2 
\end{equation}
The training and test data were collected randomly in the range \([-2, 2]\) of \(\boldsymbol{z}\). The training set contained 100 examples and the test set contained 20 examples. \vspace*{-2mm}

\noindent The structure of the fully connected network was as follows. \vspace*{-3mm}
\begin{itemize}
	\item A 2-hidden layer network with structure 2-50-50-1 (2 inputs, 50 and 50 units in hidden layers, 1 output). The activation functions at hidden layers  are modified softplus \(f(x)=log(0.8+e^x)\) and those for the output layer are identity \(f(x)=x\). \vspace*{-2mm}
\end{itemize}
With the structure of 2-50-50-1, the full network was trained {by FPL method} through training two smaller SLFNs sequentially: 2-50-1 (Subnet I), 50-50-1 (Subnet II). Each subnet was trained in 2 phases: pre-training with 1000 loops and fine-tuning with 50000 loops. \vspace*{-2mm}

The results for the training and test sets are shown in Fig. \ref{fig:sinexp_train} and Fig. \ref{fig:sinexp_test}. The mean square errors (MSEs) for these training and test sets are 1.05e-04 and 3.62e-04 respectively. \vspace*{-2mm}
\begin{figure}[ht]
	\begin{subfigure}{0.45\linewidth}
		\centering
		\includegraphics[width=1.8in]{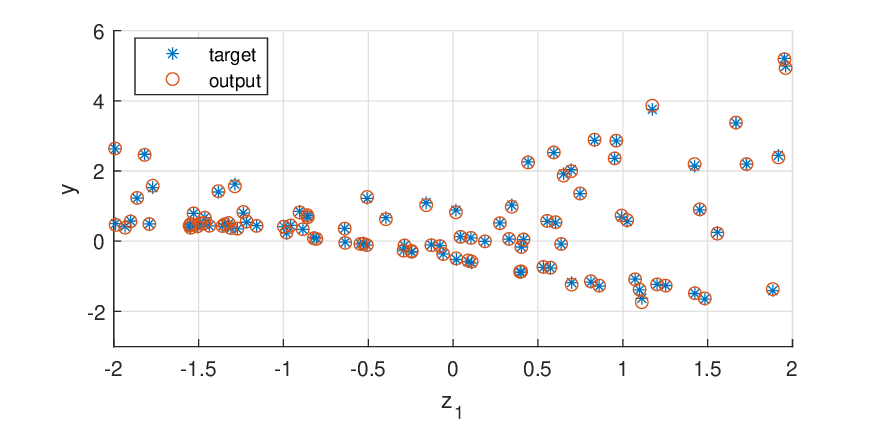} 
		\caption{Training set}
		\label{fig:sinexp_train}
	\end{subfigure}
	\begin{subfigure}{0.45\linewidth}
		\centering
		\includegraphics[width=1.8in]{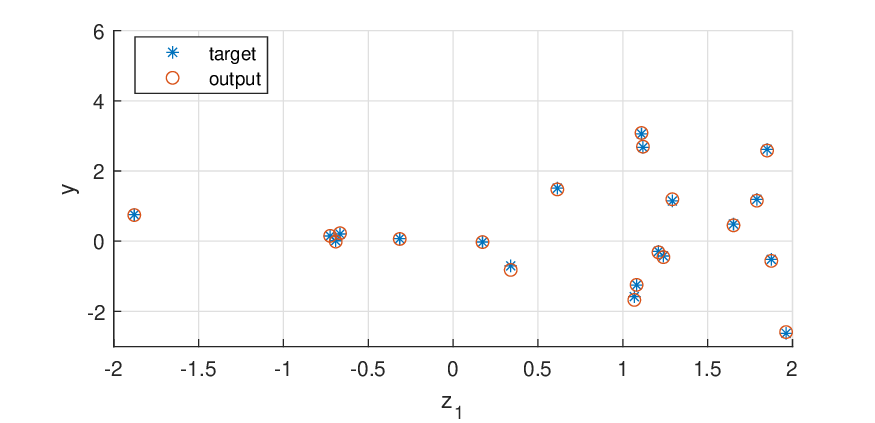} 
		\caption{Test set}
		\label{fig:sinexp_test}
	\end{subfigure}
	\caption{Results for the training set and test set in the function approximation task.}
	\vspace{-3mm}
\end{figure}
}

\subsection{MNIST Database} \label{sec: FPL_mnist} \vspace{-2mm}
We considered the same network structure as in subsection \ref{sec: mnist_Ex1}, which was 784-300-100-50-10. {Inspired by some of the configurations used in \cite{lecun1998gradient,lecun1998mnist} for the same classification task, we had tested a few network structures and finally settled with the above architecture which was sufficiently deep with reasonable number of neurons.} We call it the full network. The activation functions used at hidden layers of this full network were ReLU \(f(x)=\max(0, x)\) and those at output layer were sigmoid \(f(x)={1}/{(1+e^{-x})}\). \vspace*{-2mm}

With the structure of 784-300-100-50-10, the full network was trained through training three smaller single layer feedforward networks (SLFNs) sequentially: 784-300-10 (Subnet I), 300-100-10 (Subnet II), and 100-50-10 (Subnet III). The activation functions used at the hidden layer of all three nets were the same as those at the hidden layers of the full network (ReLU), and the output activation functions of the three nets were also the same as those of the full network (sigmoid). Each SLFN was trained in 2 phases: pre-training using one-layer update and fine-tuning using two-layer update. In the pre-training phase of each net, the identity output was used, and in the fine-tuning phase, the sigmoid output was used. After Subnet I (784-300-10) had been fully trained, its input weights (the layer 783-300) were kept and frozen, while its output weights (300-10) were discarded. This process creates a modified input layer which has a dimension of 300. This new input was fed into Subnet II (300-100-10) and the training process of Subnet II began. Similarly, after Subnet II had been trained, a modified input layer of dimension 100 was created. Training of Subnet III (100-50-10) took place afterwards. \vspace*{-5mm}

The choice of gain matrices in pre-training phase of the three nets could be done similarly to the subsection \ref{sec: mnist_Ex1}, which is by calculating from the condition in \eqref{eq:mlfn_wn_con_L}. In fine-tuning phase, the gain matrix should satisfy the condition stated in \eqref{eq:con_L_2_final}. This condition seems to suggest that the training at the ending or fine-tuning stage should be done with a small gain so as to ensure convergence. In practice, the gain was initially set to some value, and then adjusted automatically by monitoring the condition in each update and reducing it if necessary.\vspace*{-3mm}

\begin{table}[t]
	\normalsize
	\centering
	\caption{Convergence of FPL and SGD on Subnet I with different gains}
	\label{tbl: mnist_convergence}
	\setlength\extrarowheight{-2pt}
	\begin{tabular}{l|c|c}
		{\textbf{Gains }}&\textbf{FPL} & \textbf{SGD}\\
		\hline
		1000, 100, 10 & converge & diverge \\						
		\hline
		0.1, 0.01, 0.001 & converge & converge\\
		\hline
	\end{tabular}
\hspace{-2mm}	
\end{table}

\begin{table}[t]
	\normalsize
	\begin{center}
		\caption{Training \& test accuracies (\%) by FPL and SGD for MNIST dataset}
		\label{tbl: FPL_vs_SGD_sigmoid}
		\setlength\extrarowheight{-2.5pt}
		\begin{tabular}{l|c|c|c|r}
			\textbf{ }&\multicolumn{2}{c|}{\textbf{FPL}} & \multicolumn{2}{c}{\textbf{SGD}}\\
			\hline
			&training&test&training&test\\
			\hline		
			\multirow{5}*{\begin{tabular}[x]{@{}c@{}} Running\\ 5 times\end{tabular}}  &99.94 &98.59 &99.81& 98.47\\		
			&99.93 & 98.37 &99.81&98.30\\
			&99.91 & 98.43 &99.78&98.38\\
			&99.92 & 98.38 &99.81&98.26\\
			&99.89 & 98.42 &99.78&98.41\\		
			\hline
			{Mean values} &99.92 &98.44 &99.80& 98.36 \\		
		\end{tabular}
	\end{center}
\vspace{-2mm}
\end{table}

The proposed FPL algorithm was compared with the SGD method where the network was trained for all layers together. 
We first tested the convergence of the FPL and SGD algorithms on an SLFN whose structure was the same as Subnet I above by using various different gains (matrix \(\mathbf{L}\) in FPL and learning rate in SGD). Since at this stage FPL iterates for 1 training example at a time, the batch size in the case of SGD was also set as 1 for consistency. Table \ref{tbl: mnist_convergence} shows a summary of the results. Noting that for FPL, the gain matrix was for initial setting only since it can be adjusted automatically by checking the condition \eqref{eq:con_L_2_final} during the training process. It is seen from the table that FPL can guarantee the convergence for a wide range of learning gain, while divergence can occur in SGD in cases where the gain is large. Another case study on online kinematic control of robot is presented in subsection \ref{sec:sgd_fpl_simulator} to illustrate the performance of FPL as compared to SGD in dealing with new tracking tasks or new circumstances. \vspace*{-2mm}

After testing the convergence, we compared the training and test accuracies of the two methods. The learning rate (LR) for SGD was chosen small enough such that the loss function converged. The LR was initially set at 0.05, the number of epochs was 100, and the batch size was 1. The LR was reduced by 2 after half of number of epochs and by 4 after \(3/4\) of number of epochs.  For the FPL, the number of loops for training of the last layer in pre-training phase was 2. In the fine-tuning phases, the initial setting of the gain matrix for all nets was \(\mathbf{L} = \text{diag}(0.01, \cdots, 0.01)\). For each of the three nets, it took 28 loops with the gain scheduled to be decreasing when the loop number increased. To evaluate the effectiveness and consistency of the proposed method with different initial weights and random feed of training examples, we ran 5 times for each of FPL and SGD methods. The results in the 5 runs did not deviate much for both SGD and FPL. In each time, the accuracy for the test set was the maximum value (of the whole process with SGD, and of the process of training Subnet III with FPL), and the accuracy for the training set was chosen at the epoch where the test accuracy peaked. We then took average of the five values to get the mean training and test accuracies. {Similar to the practice in the experiment in subsection \ref{sec: mnist_Ex1}, since we wanted to observe the best achievable prediction for each method, no validation set was used. Instead, both SGD and FPL methods were evaluated in the same manner using the test set.}\vspace*{-2mm}

The accuracies on training set and test set of the two methods are  given in Table \ref{tbl: FPL_vs_SGD_sigmoid}. It can be seen that the final test results of the FPL algorithm are quite similar to the corresponding results obtained by the SGD method.\vspace*{-2mm}

\subsection{CIFAR-10 Database} \label{sec: FPL_cifar10} \vspace{-2mm}
The next classification task is based on CIFAR-10 database {\cite{krizhevsky2014cifar}}. \vspace{-2mm}
{\begin{itemize}
		\item CIFAR-10 is a database that contains color images of objects in 10 classes. It has 50,000 examples in the training set and 10,000 examples in the test set. {Each of these} images has the size of 32\(\times\)32 pixels.
		\vspace{-2mm}
\end{itemize} }
For CIFAR-10 dataset, we used a pre-trained convolutional neural network (ResNet-18) to get the output of the convolutional part. This is one of the common techniques in transfer learning where the convolutional layers are fixed as feature extractor, and only the classifier layers are trained for the specific task. The output or extracted features were then considered as the input of the classifier part which was a network with several layers. The output of the convolutional part had a dimension of 512. The structure of the fully connected network was as follows. \vspace{-3mm}
\begin{itemize}
	\item A 2-hidden layer network with structure 512-200-80-10 (200 units, 80 units in hidden layers). The activation functions at hidden layers are ReLU \(f(x)=\max(0, x)\) and at output layer are sigmoid \(f(x)={1}/{(1+e^{-x})}\). \vspace*{-3mm}
\end{itemize}
With the structure of 512-200-80-10, the full network was trained through training two smaller SLFNs sequentially: 512-200-10 (Subnet I), 200-80-10 (Subnet II). \vspace*{-2mm}

The training and test accuracies of the proposed method are compared with the SGD method. The SGD trains all layers of each network all together, with batch size set to be 1. The SGD method took 300 full epochs and LR = 0.01 initially. For the FPL algorithm, in the pre-training phase of the two nets, the number of loops for training of the last layer was 2. In the fine-tuning phases, the initial setting of the gain matrix for Subnet I was \(\mathbf{L} = \text{diag}(0.002, \cdots, 0.002)\) and for Subnet II was \(\mathbf{L} = \text{diag}(0.0005, \cdots, 0.0005)\). It took 128 loops for fine-tuning of Subnet I and 68 loops for Subnet II, and the gains were scheduled to be decreasing when the loop number increased. For each of FPL and SGD methods, we ran {10} times. In each time, the accuracy for the test set was the maximum value (of the {entire} process with SGD, and of the process of training Subnet II with FPL), and the accuracy for the training set was chosen at the epoch where the test accuracy peaked. We then took average of the five values to get the mean training and test accuracies. \vspace*{-2mm}

The accuracies on training set and test set of the two methods are given in Table \ref{tbl: cifar10 FPL_vs_SGD_simgoid}. We can see that the final test results of FPL method are quite similar to the corresponding results obtained by SGD method. {The slightly lower mean training accuracy {with similar test accuracy} for FPL indicates a {marginally} better generalization capability of our method over SGD in this particular problem.} \vspace*{-2mm}

{\textbf{Remark 7}: {The choice of LR is highly dependent on the dataset as well as the training conditions \cite{bengio2012practical,patterson2017deep}. It is one of the hyper-parameters that needs to be chosen with care since a high LR can lead to divergence while a low LR would slow down the training process \cite{patterson2017deep,ruder2016overview}.} The initial LRs used in the experiments of SGD were chosen from several empirical trials to avoid divergence and to achieve the highest accuracies for each dataset. As for the proposed FPL, convergence is no longer an issue when selecting an initial gain since the gain can be reduced automatically during training by checking the condition given in \eqref{eq:con_L_2_final}. {Nevertheless,} choosing a suitable initial gain would {still} benefit the training process in terms of training time and the final results. {In order to observe the best possible outcome for all compared methods,} we have set the best values for the initial gains for both SGD and FPL such that the highest {test} accuracies have been achieved.} \vspace*{-2mm}
\begin{table}[t]
	\captionsetup{labelfont={color=blue}}
	\normalsize
	\begin{center}
		\caption{Training \& test accuracies (\%) in FPL and SGD for CIFAR-10 dataset}
		\label{tbl: cifar10 FPL_vs_SGD_simgoid}
		\setlength\extrarowheight{-2.5pt}
		\begin{tabular}{l|c|c|c|r}
			\textbf{ }&\multicolumn{2}{c|}{\textbf{FPL}} & \multicolumn{2}{c}{\textbf{SGD}}\\
			\hline
			&training&test&training&test\\
			\hline		
			\multirow{10}*{\begin{tabular}[x]{@{}c@{}} Running\\ {10} times\end{tabular}} 
			&94.49 &88.24 &94.41& 88.20\\		
			&94.14&88.22 &94.85& 88.17\\
			&93.98 &88.12 &94.17& 88.14\\
			&94.17 &88.19 &94.62& 88.12\\
			&94.41 &88.19 &94.78& 88.20\\
			&{94.41} &{88.32} &{94.71} & {88.17} \\	
			&{94.28} &{88.15} &{94.61} &{88.20}  \\
			&{93.99} &{88.14} &{94.41} & {88.24} \\
			&{94.39} &{88.26} &{94.74} & {88.16} \\
			&{94.46} &{88.17} &{94.71} & {88.21} \\	
			\hline
			{Mean values} &{94.27} &{88.20} &{94.60}& {88.18} \\		
		\end{tabular}
	\end{center}
	\vspace{-2mm}
\end{table}

\subsection{Online Kinematic Control: UR5e Robot} \label{sec: ur5e} \vspace{-2mm}
{Although it is good to ensure convergence of the deep learning systems in classifications problems so as to establish a systematic method instead of trial and error method for selection of learning gains, it may be sometimes arguable that the convergence analysis is not very critical in classification problems since there is no harm to redo the training if divergence occurs. However, for online training of robots, convergence is crucial in assuring a safe operation at all time. In this section, we firstly show the importance of convergence in the online learning in robot control by comparing the performance of SGD and FPL on the simulator. After that, we show how a deep network can be built progressively by using FPL method such that the convergence of the online feedback error is guaranteed on the real robot.} By repeating the operations, we shall show that the robot can gradually learn to execute a task based on feedback errors of the end effector without any knowledge of the kinematic model. \vspace*{-3mm}

The tracking control task was drawing a circle in 3D space. The desired trajectory in sensory space is a circle (C1) specified as
\begin{eqnarray}
x_1 &=& -0.4 - 0.06 \cos\left(\omega t \right)  -0.18 \sin\left(\omega t \right) \nonumber\\
x_2 &=& 0.2 \cos\left(\omega t \right)  -0.02 \sin\left(\omega t \right) \nonumber\\
x_3 &=& 0.5 + 0.02 \cos\left(\omega t \right) + 0.06 \sin\left(\omega t \right) \label{eq:desire_traj} 
\end{eqnarray}
The units for the coordinates are meter (m). The circle (C1) has a center at \([-0.4, 0, 0.5]\) and a radius of 0.2 m. 

\subsubsection{Convergence tests of SGD and FPL on the simulator}\label{sec:sgd_fpl_simulator} \vspace{-2mm}
Because of safety reason, we only tested SGD on the simulator as there is no guarantee of convergence when using SGD in online learning. For the purpose of comparison, FPL was also tested on the same simulator, before it was finally implemented on the actual robot (in subsection \ref{sec:fpl_real}).  \vspace*{-2mm}

We considered the major axis which includes the first 3 joints \(q_1,q_2,q_3\). A single hidden layer network was built to approximate the Jacobian matrix of the UR5e robot through the relationship in \eqref{eq:jacob_all}. To do that, we firstly learned an SLFN with structure 3-12-3. That is, 3 input nodes (for \(q_1,q_2,q_3\)), 12 nodes in the hidden layer and 3 output nodes (for \(\dot{x}_1,\dot{x}_2,\dot{x}_3\)). The activation functions used for the hidden layers were modified softplus \(f(x)=log(0.8+e^x)\) and for the output layer were identity \(f(x)=x\).  \vspace*{-2mm}

Since the kinematic model is unknown, we needed to first manually 
move the robot around the desired trajectory in order to collect data for offline training of the network. The data of \(\boldsymbol{q}\), \(\dot{\boldsymbol{q}}\) and \(\dot{\boldsymbol{x}}\) were collected during the manual movement. After getting the data, we trained the network offline by using SGD method. Different from the convergence test in the classification task, the learning rate here was chosen such that the offline learning converged.
The obtained weights were then adopted as a starting point for the learning of Jacobian matrix in the online training. \vspace*{-2mm}

We performed the online training using both SGD and FPL on the simulator of UR5e. The robot was first moved to an initial position so that the initial error was zero. The training was then conducted by using the online \(\dot{\boldsymbol{q}}\) command as constructed in \eqref{eq:control_qre}.
\begin{equation}
\dot{\boldsymbol{q}}(k) = \hat{\mathbf{J}}^\dagger(\boldsymbol{q}(k),\hat{\mathbf{W}}_{\Sigma}(k))(\dot{\boldsymbol{x}}_d(k) -\alpha\Delta \boldsymbol{x}(k)) \nonumber
\end{equation}
The weights of the network were subsequently updated online using the SGD with estimation error \(\dot{{\boldsymbol{x}}}(k) - \dot{\hat{\boldsymbol{x}}}(k)\) and the FPL with online feedback error \(\boldsymbol{\varepsilon}(k) \) in \eqref{eq:dyna_erro}. The matrix \(\hat{\mathbf{J}}(\boldsymbol{q}(k),\hat{\mathbf{W}}_{\Sigma}(k))\) can be calculated based on current joint variables \(\boldsymbol{q}(k)\) and current weights \(\hat{\mathbf{W}}_{\Sigma}(k)\). With both methods, the same gain (learning rate) as used in the offline learning phase was used. To test the performance of the robot system in tracking new task, the desired speed of the end effector was increased (3-5 times) as compared to the original speed when moving the robot manually. \vspace*{-2mm}

\begin{figure}[t]
	\centering
	\includegraphics[width=3.3in]{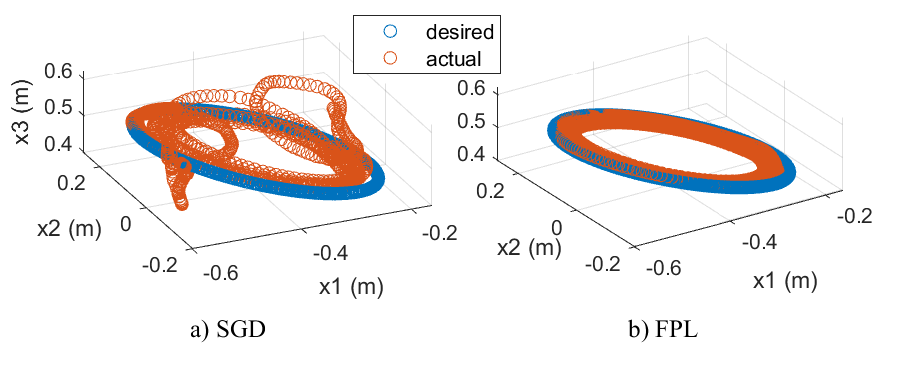} \vspace{-3mm}
	\caption{Desired and actual trajectories of the robot end-effector in the convergence test of online learning using SGD and FPL methods: a) SGD - Divergence occurred even though the same learning rate as in offline learning phase was used; b) FPL - Convergence was guaranteed by using the proposed online adjustment of gain matrix.}
	\label{fig:sgd}
	\vspace{-1mm}
\end{figure}

Fig. \ref{fig:sgd}a shows the plot of the actual path of the robot end effector by using SGD to approximate the Jacobian matrix. It can be seen that the end effector deviates significantly from the desired path during the on-line training. The program terminated after some time, as the learning of the network became unstable causing interruption in the calculation of \(\hat{\mathbf{J}}^\dagger(\boldsymbol{q}(k),\hat{\mathbf{W}}_{\Sigma}(k))\). The result in Fig. \ref{fig:sgd}b shows that the convergence can be ensured by using FPL to achieve safe on-line learning.  \vspace*{-2mm}

\subsubsection{FPL on the real robot} \label{sec:fpl_real} \vspace{-2mm}
With the real robot, the real values of joint variables \(\boldsymbol{q}\) and joint velocities \(\dot{\boldsymbol{q}}\) were collected using the internal communication channel of the robot. The positions \(\boldsymbol{x}\) of the end effector in sensory space were recorded using a Kinect RGB-D camera. {The precision of the RGB-D {camera} is approximately 2 mm for {measuring} distance {between} 50 cm to less than 1 m, {which was the operating range of} the robot in our experiments.}  The sampling time was about 0.07 s. {The velocities \(\dot{\boldsymbol{x}}\) were then calculated by differentiating the positions \(\boldsymbol{x}\).} \vspace{-2mm}

{For the experiment on real robot, a setpoint control task and a trajectory tracking control task {were defined}. For the setpoint control task, the	robot end-effector {was commanded} to move to specific positions in the task space. The setpoint was changed after an interval of every 30 {seconds}. For the trajectory tracking task, the desired trajectory was the circle (C1) specified by \eqref{eq:desire_traj}.} The angular frequency (or angular speed) \(\omega\) in \eqref{eq:desire_traj} was planned in 5 phases. In the first 5 seconds, the desired \(\omega\) was 0 rad/s, which aimed to keep the robot at the initial position. The next 30 seconds (5 s - 35 s) was the acceleration period, when the desired angular frequency increased gradually from rest at 0 rad/s to full speed at \(2\pi/30\) rad/s (2 rpm). After that, the robot end effector would move 3 rounds (revolutions) at full speed in 90 seconds (35 s - 125 s), before decelerating from full speed to 0 rad/s in the next 30 seconds (125 s - 155 s). Finally, the robot would be at rest for the last 5 seconds. \vspace*{-2mm}

{\textbf{Building and training the network:}
We aimed to build a two-hidden layer network with structure 3-12-12-3 to approximate the Jacobian matrix of the UR5e robot through the relationship in \eqref{eq:jacob_all}. To do that, we firstly learned an SLFN with structure 3-12-3 (Subnet I) (similar to the network used in the simulator above). After Subnet I had been trained, we discarded its output weights and then added one new hidden layer with 12 neurons. Because the input weights of {the newly formed} network were frozen, the training process was continued with a network of structure 12-12-3 (Subnet II). The 2 subnets were trained sequentially and not alternately. That is, Subnet I was trained first. After Subnet I had been trained in the online task, there was no more training of Subnet I taking place. Subnet II was then built and trained in a similar way as Subnet I. The training process ended after Subnet II had been trained in the online task. \vspace*{-2mm}}

\textbf{Training of the first hidden layer network (Subnet I):}
Though the proposed online kinematic control algorithm in section \ref{sec: main_3} can be applied directly, the transient performance at the initial stage of learning may not be good since the controller is completely model-free at the initial stage.
To overcome this issue, we adopt a combination of offline and online trainings so that real-time feedback control using deep networks can be established eventually. \vspace*{-2.5mm}

After getting the manual data, we first trained the network offline by one-layer pre-training (as in subsection \ref{sec:pre-train}) and two-layer fine-tuning (as in subsection \ref{sec:fine_tuning}). The obtained weights were then adopted as a starting point for the learning of Jacobian matrix in the online training. \vspace*{-2.5mm}

{ {For online training, the robot was first moved to an initial position near the first setpoint for the {task of} setpoint control, and to an initial position so that the initial error was zero for the trajectory control task}. The training was then conducted by using the online \(\dot{\boldsymbol{q}}\) command as constructed in \eqref{eq:control_qre}. 
	The weights of the network were subsequently updated online using the two-layer update with error \(\boldsymbol{\varepsilon}(k)\). The matrix \(\hat{\mathbf{J}}(\boldsymbol{q}(k),\hat{\mathbf{W}}_{\Sigma}(k))\) can be calculated based on current joint variables \(\boldsymbol{q}(k)\) and current weights \(\hat{\mathbf{W}}_{\Sigma}(k)\). } \vspace{-2.5mm}
\begin{figure}[ht]
	\begin{subfigure}{0.48\linewidth}
		\includegraphics[width=1.65in]{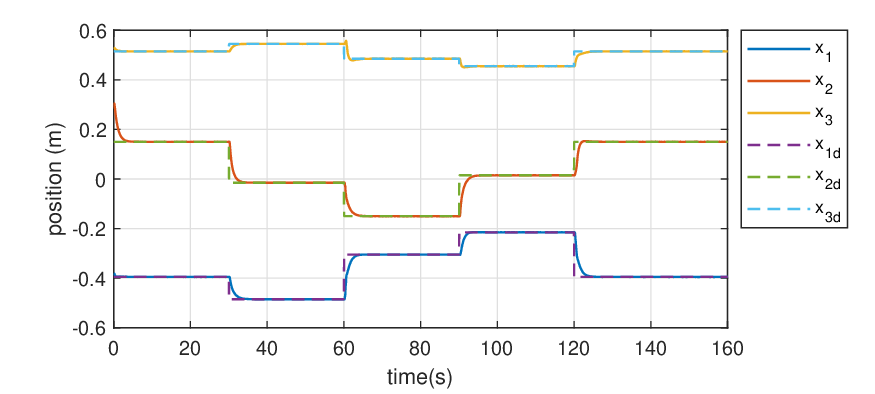} 
		\caption{Training setpoints}
		\label{fig:1}
	\end{subfigure}
	\begin{subfigure}{0.48\linewidth}
		\centering
		\includegraphics[width=1.65in]{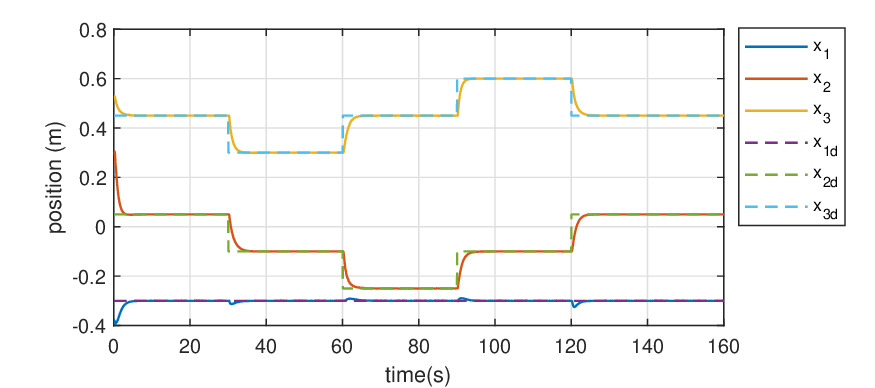} 
		\caption{Test setpoints}
		\label{fig:2}
	\end{subfigure}
	\caption{The actual (solid lines) and desired (dashed lines) positions of robot end-effector in setpoint control tasks.}
	\vspace{-3mm}
\end{figure}
\begin{figure}[t]
	\centering
	\includegraphics[width=2.2in]{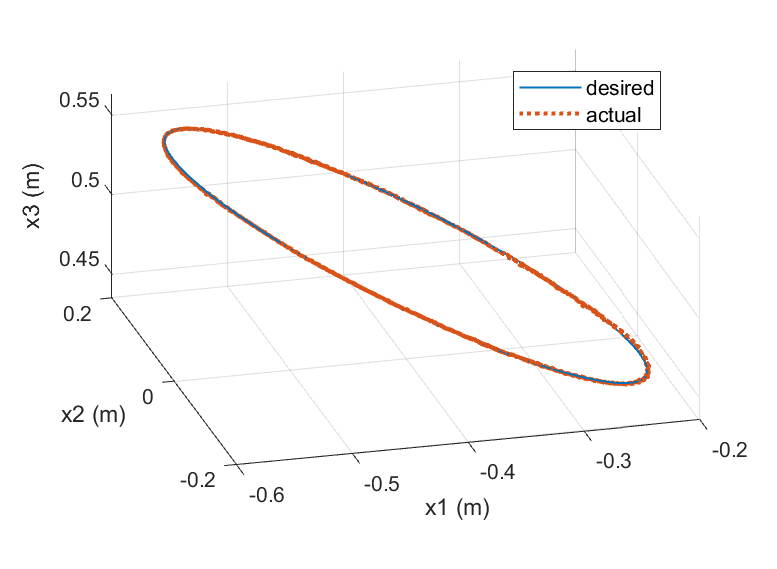}
	\caption{Desired and actual trajectories of the robot end-effector in sensory space for training cicle (C1).}
	\label{fig:final_traj}
	\vspace{-2mm}
\end{figure}
\begin{figure*}[t]
	\centering
	\includegraphics[width=6.2in]{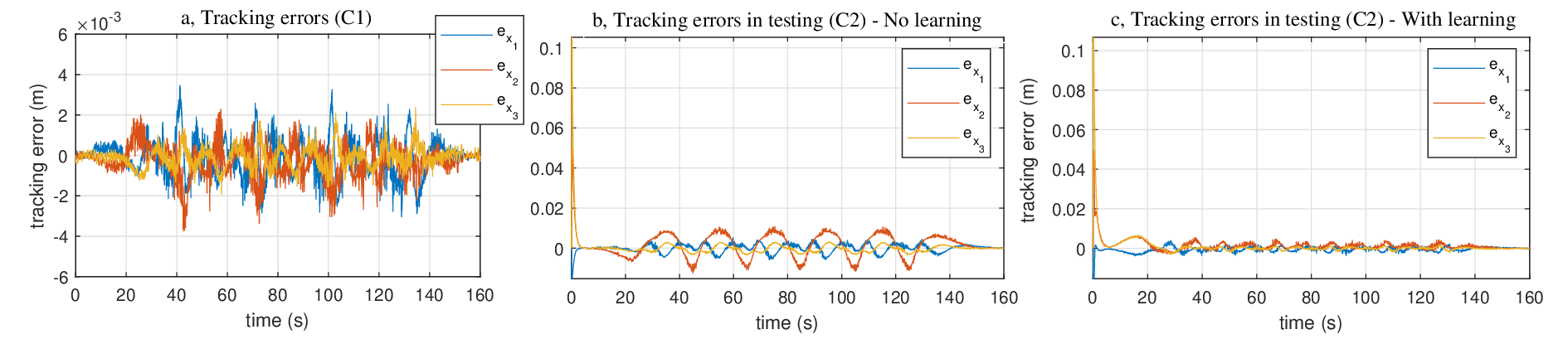}
	\vspace{-2mm}
	\caption{Performance of the online kinematic control task on the real UR5e robot. The first figure is for the training circle (C1):  a, tracking error (with respect to time) of every coordinate.  The last 2 figures are for the testing circle (C2): b, tracking errors when the weights of the network are fixed and c, tracking errors when the last layer of the network is updated.}
	\label{fig:1stnet_error}
	\vspace{-2mm}
\end{figure*}

{\textbf{Training of the second hidden layer network (Subnet II):}}
The learning of this net was similar to that of Subnet I. Again, offline training was first performed to avoid poor transient performance. The training data for offline training of Subnet II were a combination of half of the amount of manual data and half of the amount of new data generated in the online training of Subnet I. After offline training, the online control task was done similarly to Subnet I. \vspace*{-2mm}
	
{The actual and {the} desired positions in the setpoint control task are shown in Fig. \ref{fig:1}. For the trajectory tracking task,} the desired and actual trajectories for online learning are shown in Fig. \ref{fig:final_traj}, and the tracking error is shown in Fig. \ref{fig:1stnet_error}a. It can be observed that the errors are very small and the actual trajectory follows the desired one very closely. This tells us that building network using FPL can guarantee the convergence of the tracking errors in online learning control. \vspace*{-2mm}

\textbf{Testing of the trained network:} To test the generalization	property of the network, we used the Jacobian matrix obtained after online training of Subnet II above for a tracking control task with a new trajectory which was also a circle (C2) with radius of 0.15 m. This circle was on a new plane which is 0.1 m lower along the \(x_3\)-axis compared with (C1). The maximum speed of the movement was at \(2\pi/20\) rad/s (or 3 rpm) and the direction of the movement was opposite. {To illustrate the generalization property, we also tested the network trained {under the task of} setpoint control for a new set of setpoints that were not used in the training.}\vspace*{-2mm}
	
{Fig. \ref{fig:2} shows the control performance for the unseen setpoints in the setpoint control task.} Fig. \ref{fig:1stnet_error}b and \ref{fig:1stnet_error}c show the tracking errors for the new trajectory (C2). The initial errors for \(x_3\)-axis are large (about 0.1 m) as the robot did not start on the new desired trajectory. The initial position of the end effector was set as the same as the old trajectory (C1) used for training.
Fig. \ref{fig:1stnet_error}b shows the tracking errors when the Jacobian is used directly without any update of the weights. It can be seen that the peaks in the full-speed period (35 s - 125 s) are similar to each other (at about 0.01 m), which means that the errors are the same for each evolution of the movement of the end effector. This is understandable as the weights are kept constant during the new tracking control task. 
Fig. \ref{fig:1stnet_error}c shows the tracking errors when the Jacobian is updated during the online control. Only the weights of the last layer were trained during online learning. It is observed that the peaks in the full-speed period (35 s - 125 s) now are much smaller than the previous case. \vspace*{-2mm}
	
Hence, from this case study, we can see that the FPL framework ensures the convergence of the tracking errors in the online learning. The framework also gives quite good generalization in this experiment {in the sense that even though the test setpoints and trajectory have not been used for training the networks, the networks still perform well on these unseen test data.} And only the weights of the last layer are updated during online control task so as to achieve better tracking performance for the new trajectory.  \vspace*{-2mm}
	
\section{Conclusion} \vspace{-2mm}
In this paper, we have presented a layer-wise deep learning framework in which a multilayer {fully connected} network can be built and trained such that the convergence of the algorithm is ensured. The case studies of classification tasks using MNIST and CIFAR-10 databases have shown that using the learning framework can {yield} similar accuracy as gradient descent method while ensuring convergence. {It has also been shown that a robot can learn to execute an online kinematic control task in a safe and predictable manner without any modeling.} We believed that the proposed method would widen the potential applications of deep learning in the areas of robotics and control. \vspace*{-2.5mm}

{\color{black}{This paper considers fully connected deep neural networks which can be used for general classification and approximation problems. For image classification and pattern recognition problems, a special and important class of deep networks called convolutional neural networks (CNNs) (which include convolutional layers and fully connected layers) are found to be more effective. Future work would {thus} be devoted to developing layer-wise learning method for CNNs and analyzing {their learning} convergence.} The parameter sharing mechanism in CNNs, which apparently differs from {the} fully connected layers in MLFNs, would be an interesting topic to explore. Another possibility would be easing the requirement of pre-training in the two-layer update algorithm. {Apart from layer-wise learning, future work would also be devoted to developing {methodology to update} the weight matrices of all layers concurrently.} {From the aspect of} applications, other online robot control problems in dynamic vision based manipulators or robotic systems with different kinematic and dynamic structures such as mobile robots, multiple robot systems, etc would also be worth exploring.}
\vspace{-2.5mm}

\bibliographystyle{ieeetran}
\bibliography{autosam}           



\end{document}